\documentclass[10pt,journal,compsoc]{IEEEtran}

\usepackage{amsmath}
\usepackage{multirow}
\usepackage{booktabs}
\usepackage{color}
\usepackage{graphicx}
\usepackage{algorithmic}
\usepackage{float}
\usepackage{times}
\usepackage{epsfig}
\usepackage{amssymb}
\usepackage{color,xcolor} 
\usepackage{caption}
\usepackage{subcaption}
\usepackage{array}
\usepackage{ragged2e}
\usepackage[section]{placeins}
\usepackage{hyperref}
\usepackage{utfsym}
\usepackage{url}

\ifCLASSOPTIONcompsoc
  \usepackage[nocompress]{cite}
\else
  \usepackage{cite}
\fi

\ifCLASSINFOpdf
\else
\fi

\begin{document}
%
\title{
Pushing Auto-regressive Models for 3D Shape Generation at Capacity and Scalability
}

\author{Xuelin Qian$^\star$, Yu Wang$^\star$, Simian Luo, Yinda Zhang, Ying Tai, Zhenyu Zhang, Chengjie Wang \\ Xiangyang Xue, Bo Zhao, Tiejun Huang, Yunsheng Wu, Yanwei Fu
\thanks{$^\star$ indicates equal contributions.}

\IEEEcompsocitemizethanks{
\IEEEcompsocthanksitem 
Xuelin Qian, Yu Wang and Yanwei Fu are with the School of Data Science, and Fudan ISTBI—ZJNU Algorithm Centre for Brain-inspired Intelligence, Fudan University, China. Email: \{xlqian, yu\_w13, yanweifu\}@fudan.edu.cn. 
Xuelin Qian and Yu Wang have equal contributions. 
\IEEEcompsocthanksitem 
Simian Luo is with the Artificial Intelligence at IIIS (Institute for Interdisciplinary Information Sciences), Tsinghua University, China. Email:
luosm22@mails.tsinghua.edu.cn.
\IEEEcompsocthanksitem Yinda Zhang is with Google. Email: zhangyinda@gmail.com.
\IEEEcompsocthanksitem 
Ying Tai and Zhenyu Zhang are with the School of Intelligence Science and Technology, Nanjing University (Suzhou Campus), China. Email:
yingtai@nju.edu.cn, zhangjesse@foxmail.com.
\IEEEcompsocthanksitem 
Chengjie Wang and Yunsheng Wu are with Tencent Youtu lab, Shenzhen, China. Email: 
\{jasoncjwang, simonwu\}@tencent.com
\IEEEcompsocthanksitem 
Bo Zhao and Tiejun Huang are with BAAI (Beijing Academy of Artificial Intelligence). Email: zhaobo@baai.ac.cn, tjhuang@pku.edu.cn.
\IEEEcompsocthanksitem 
Xiangyang Xue is with the School of Computer Science, and  Shanghai Key Lab of Intelligent Information Processing, Fudan University, China. Email: xyxue@fudan.edu.cn.
}}


\IEEEtitleabstractindextext{%
\begin{abstract}
Auto-regressive models have achieved impressive results in 2D image generation by modeling joint distributions in grid space. In this paper, we extend auto-regressive models to 3D domains, and seek a stronger ability of 3D shape generation by improving auto-regressive models at capacity and scalability simultaneously. Firstly, we leverage an ensemble of publicly available 3D datasets to facilitate the training of large-scale models. It consists of a comprehensive collection of approximately $900,000$ objects, with multiple properties of meshes, points, voxels, rendered images, and text captions. 
This diverse \textit{labeled} dataset, termed \textit{Objaverse-Mix}, empowers our model to learn from a wide range of object variations. For data processing, we employ four machines with 64-core CPUs and 8 A100 GPUs each over four weeks, utilizing nearly 100TB of storage due to process complexity. And the dataset is on \href{https://huggingface.co/datasets/BAAI/Objaverse-MIX}{https://huggingface.co/datasets/BAAI/Objaverse-MIX}.
However, directly applying 3D auto-regression encounters critical challenges of high computational demands on volumetric grids and ambiguous auto-regressive order along grid dimensions, resulting in inferior quality of 3D shapes. To this end, we then present a novel framework \textit{Argus3D} in terms of capacity.
Concretely, our approach introduces discrete representation learning based on a latent vector instead of volumetric grids, which not only reduces computational costs but also preserves essential geometric details by learning the joint distributions in a more tractable order. The capacity of conditional generation can thus be realized by simply concatenating various conditioning inputs to the latent vector, such as point clouds, categories, images, and texts. In addition, thanks to the simplicity of our model architecture, we naturally scale up our approach to a larger model with an impressive $3.6$ billion parameters, further enhancing the quality of versatile 3D generation. 
Extensive experiments on four generation tasks demonstrate that Argus3D can synthesize diverse and faithful shapes across multiple categories, achieving remarkable performance. The dataset, codes and models are available on our project website  \href{https://argus-3d.github.io}{https://argus-3d.github.io}.
\end{abstract}

}

\maketitle

\IEEEdisplaynontitleabstractindextext

\IEEEpeerreviewmaketitle

\IEEEraisesectionheading{\section{Introduction}\label{sec:introduction}}

The 3D shape generation has garnered increasing interest in both academia and industry due to its extensive applications in robotics \cite{mees2019self}, autonomous driving \cite{ye2021online,qian2022impdet}, augmented reality \cite{sun2018x} and virtual reality \cite{stets2017visualization}. 
Based on whether user prerequisites are provided, shape generation is typically categorized as unconditional or conditional.
For a generative model to be effective, 
it is crucial for the synthesized shapes to be both \textit{diverse} and \textit{faithful}  to the universal cognition of humans or given conditions. These qualities serve as the foundation for other deterministic tasks, such as shape completion, single-view reconstruction, and more.
Previous approaches~\cite{chen2019learning,ibing20213d,park2019deepsdf} typically employ an AutoEncoder (AE) to learn latent features through shape reconstruction. 
Then, a Generative Adversarial Network (GAN) is trained to fit the distributions of these latent features, enabling the generation of 3D shapes by sampling the latent codes learned in the AE.
While these approaches yield convincing results, they still struggle with the issues of poor capacity and scalability.

Recently, the progress of Large Language Models (LLMs) \cite{thoppilan2022lamda,gpt1,gpt2} show that more parameters enable  handling complex tasks and achieving advanced performance. A natural question thereby arises: 
can we learn a comparable large model for versatile 3D shape generation?
Intuitively, we think it demands (1) a very large-scale  3D dataset ideally with corresponding labels, and (2) a quite scalable 3D model with huge parameters. 
Unfortunately, the classical  3D dataset ShapeNet~\cite{shapenet} only contains 55 categories with about 51K shapes, with relatively small data scale, and imbalanced 3D instance distribution among categories. Thus it may easily be overfitted by large 3D models.  
The newly collected very large-scale dataset, Objaverse~\cite{objaverse},  contains massive data of 800K 3D shapes. While such a dataset is a bit noisy 3D knowledge from data, and lacks of 3D annotations, Objaverse surely provides a very important data source for the improved dataset in this paper.

\begin{figure*}
\begin{centering}
\includegraphics[width=0.99\linewidth]{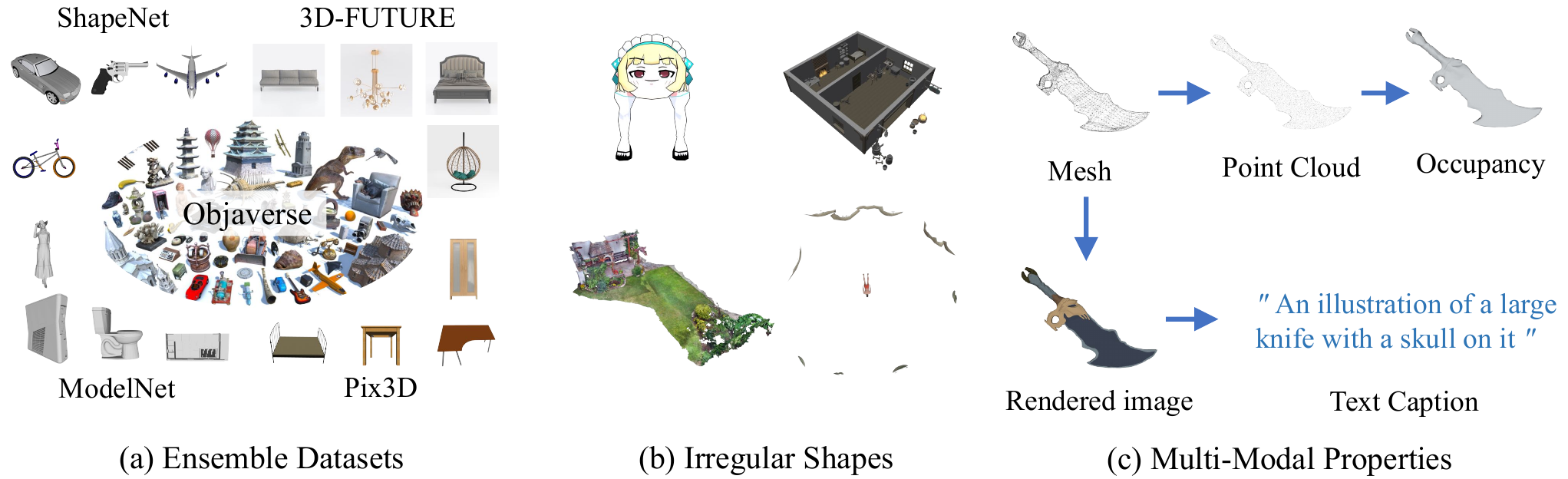}
\tabularnewline
\vspace{-0.1in}
\caption{(a) We have combined five public 3D shape datasets, amassing a total of approximately 900,000 diverse shapes. (b) We manually filter out some noisy shapes, such as irregular shapes, complex scenes, non-watertight meshes and discrete shapes. (c) Our Objaverse-Mix dataset includes meshes, point clouds, occupancies, rendered images, and text captions, showcasing its multi-modal properties.
\label{fig:datateaser} }
\end{centering}
\end{figure*}

In this paper, our first contribution is to build a large-scale 3D dataset, termed \textit{Objaverse-Mix}. Concretely, we harness an ensemble of publicly-available 3D datasets, such as ModelNet40 \cite{modelnet}, ShapeNet \cite{shapenet}, Objaverse \cite{objaverse} \textit{etc.} 
This not only allows us to efficiently scale the data, but also requires almost no set-up cost. Despite many existing 3D datasets are existed, we notice one significant drawback is the difference in quality and diverse 3D instance distribution of categories in these 3D datasets.
To this end, we systematically present the pre-processing strategies to improve the quality of our 3D dataset. Particularly, we start by normalizing and re-scaling all shapes, such that all samples have a uniform data standard. Subsequently, we adopt the off-the-shelf tools to convert 3D shapes to a unified mesh or voxel representation, and generate multiple 3D labels, such as occupancy.
Additionally, we prune shapes that do not have watertight meshes, and filter out noisy samples (\textit{e.g.}, with weird shapes, cluttered backgrounds or multiple objects), as illustrated in Fig.~\ref{fig:datateaser}b.
Critically, we utilize four machines, each with a 64-core CPU and 8 A100 GPUs, running over four weeks, and consuming nearly 100TB of storage due to process complexity.
As a result, our Objaverse-Mix comprises an impressive collection of nearly $900,000$ objects with various properties of meshes, points, voxels, rendered images and captions. Examples are shown in Fig.~\ref{fig:datateaser}c.
This diverse dataset empowers our model to learn from a wide range of object variations, promoting the visual quality of the generated 3D shapes.

With such a cleaned dataset, we turn to the recent hardware lottery structure, \textit{i.e.}, Auto-Regressive (AR) models, for  pursuing the capable 3D models. AR models have shown remarkable performance in generating 2D images~\cite{esser2021taming,zhao2021improved,chang2022maskgit} and 3D shape~\cite{mittal2022autosdf,yan2022shapeformer}. 
Instead of learning a continuous latent space, these models leverage discrete representation learning to encode each 2D/3D input into grid-based discrete codes.
Subsequently, a transformer-based network is employed to jointly model the distribution of all codes, which essentially reflects the underlying prior of objects, facilitating high-quality generation and tractable training.
However, directly applying AR models to 3D still suffers from two limitations. 
First, as the number of discrete codes increases exponentially (from squared to cubed), the computational burden of the transformer grows dramatically, making it difficult to be converged. 
Second, discrete codes in the grid space are highly coupled. It is ambiguous to simply flatten them for auto-regression,  \textit{e.g.}, with a top-down row-major order. This may lead to poor quality or even collapse of generated shapes (see Sec.~\ref{subsec:preliminary} for more discussion).

\begin{figure*}
\begin{centering}
\includegraphics[width=0.99\linewidth]{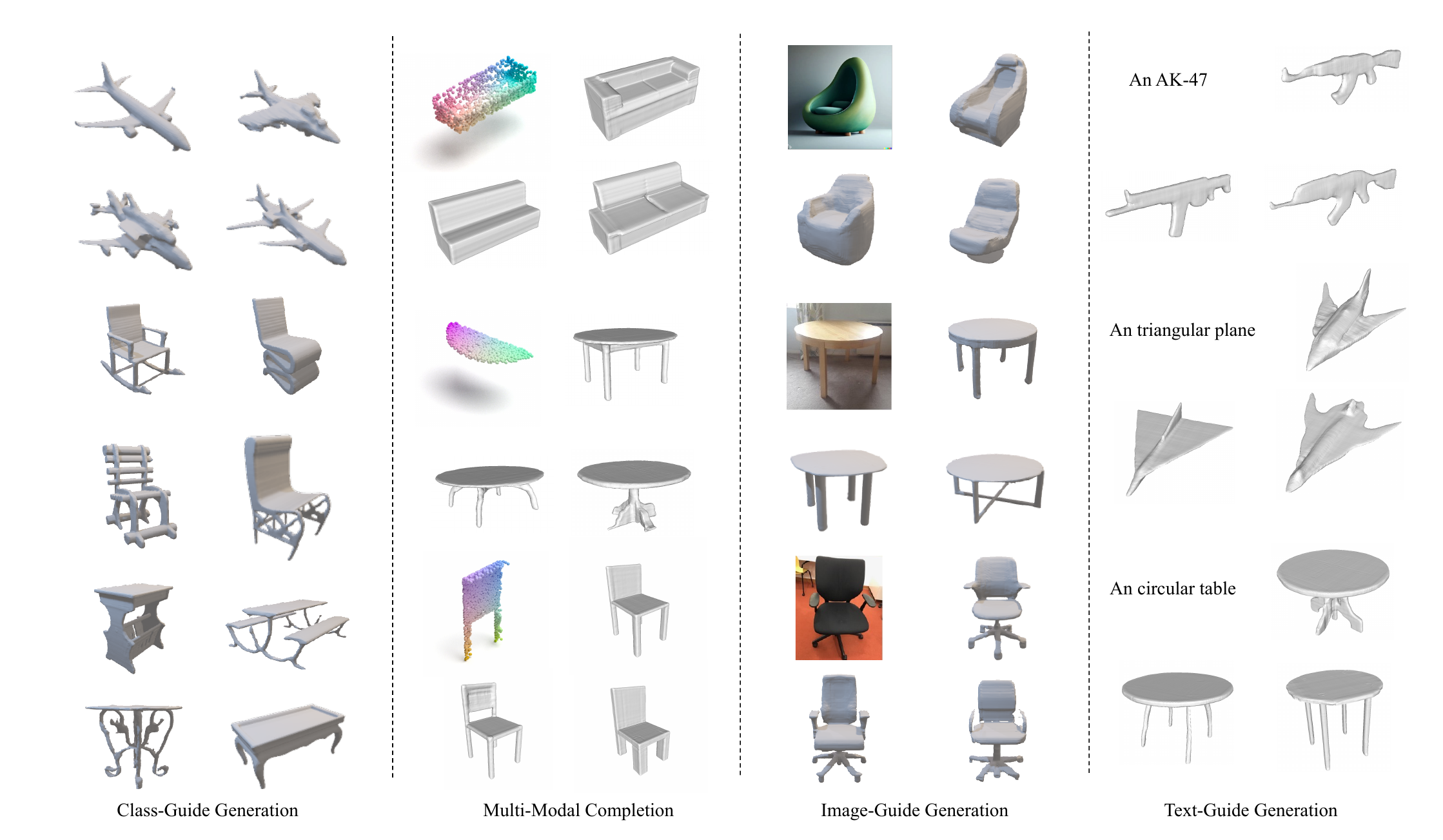}
\tabularnewline
\vspace{-0.05in}
\caption{We propose an improved auto-regressive model to learn versatile 3D shape generation. Our approach can either generate diverse and faithful shapes with multiple categories via an unconditional way (one column to the left), or can be adapted for conditional generation by incorporating various conditioning inputs given on the left-top (three columns to the right).\label{fig:teaser}
}
\end{centering}
\end{figure*}

Thus our second contribution is to propose an improved auto-regressive model in terms of \textit{capacity}, to enhance the efficient learning of 3D shape generation. Our key idea is to apply discrete representation learning in a one-dimensional space, as opposed to a 3D volumetric space. Specifically, we first project volumetric grids encoded from 3D shapes onto three axis-aligned orthogonal planes. 
This process significantly reduces the computational costs from cubic to quadratic levels while maintaining the essential information about the input geometry. 
Next, we present a coupling network to further encode three planes into a compact and tractable latent space, on which discrete representation learning is performed.
Our design is straightforward and effective, simply addressing the aforementioned limitations through two projections. 
Thus, a vanilla decoder-only transformer can be attached to model the joint distributions of codes from the latent spaces. 
We are further capable of switching freely between unconditional and conditional generation by concatenating various conditioning inputs to the latent codes, such as point clouds, categories, images and texts. 
Figure~\ref{fig:teaser} illustrates the capacity of our improved model to generate diverse and accurate shapes across multiple categories, both with and without the given conditions on the top-left corner.
Critically, inspired by the success of GPT-3~\cite{gpt3}, we naturally scale up transformers to further strengthen our auto-regressive architecture. It is implemented by increasing the number of layers and feature dimensions. Consequently, our endeavor establishes a novel 3D model with remarkable $3.6$B parameters, which further enhances the quality of versatile 3D generation as validated in Sec.~\ref{sec:scale}.

Overall, we push auto-regressive models for 3D shape generation at both capacity and scale. To this end, we propose \textit{Argus3D}, a novel framework that not only eliminates the aforementioned limitations of applying auto-regressive models to the field of 3D generation, but also scales up the model to an unprecedented magnitude of $3.6$ billion parameters, 
supporting versatile 3D shape generation under multi-modal conditions.

\noindent \textbf{Contributions.} We summarize the key contributions as follows,
(1) We propose a novel auto-regressive framework, termed \textit{Argus3D}, for 3D shape generation. Our Argus3D enjoys the advantages of switching between unconditional and conditional generation for a variety of conditioning inputs, including point clouds, categories, images, and texts.
(2) We introduce an improved discrete representation learning, 
which builds joint distributions for a latent vector instead of volumetric grids or triple planes. 
It develops auto-regressive models for 3D shape generation in terms of capacity, by tackling two limitations of high computational demands on 3D grids and ambiguous order for auto-regression.
(3) We further promote the generation ability of our Argus3D in terms of scalability, by expanding the model parameters and dataset size. To the best of our knowledge, Argus3D is the largest 3D generation model, which contains $3.6$ billion parameters and is trained on approximately $900,000$ 3D shapes across various categories. Our collected dataset, named \textit{Objaverse-Mix}, assembles a series of publicly-available 3D datasets and includes multiple properties after pre-processing, hoping to encourage more researches and advance the field of 3D shape generation.
(4) Extensive experiments are conducted on four tasks to demonstrate that our Argus3D can generate more faithful and diverse shapes, achieving remarkable performance for unconditional and conditional shape generation. 

\noindent \textbf{Extensions.} A preliminary version of this work was published as ImAM \cite{iccv}. In this paper, we highlight that our Argus3D has made several significant improvements over. More concretely,
(1) We strengthen the generation ability of our auto-regressive model by expanding its scale. We systematically design and increase the number of transformer decoder layers and feature dimensions, in order to establish the large-scale 3D shape generation model with $3.6$ billion parameters.
(2) To facilitate the training of large models, we expand the scale of datasets and provide a large-scale 3D shape dataset with approximately $900,000$ objects, which is built by assembling multiple publicly-available 3D repositories. We  filter and pre-process the data to endow each reliable 3D shape with rich attributes, such as meshes, point clouds, voxels and rendered images, to promote the research of 3D generation.
(3) We evaluate our proposed method with more distance metrics, including Chamfer distance and Earth mover's distance, further demonstrating the superiority of our improved Argus3D in 3D shape generation.
(4) We conduct extensive ablation studies to demonstrate the effectiveness of our proposed Argus3D, providing an in-depth analysis of our improvements with respect to the capacity and scale. 
(5) We achieve new state-of-the-art performance on 
multiple 3D shape generation tasks, synthesizing more diverse and faithful shapes across multiple categories.

\section{Related work}

\noindent \textbf{3D Generative Models}.
As an extremely challenging task, 
most previous efforts are made by using voxel, point clouds, and implicit representations:
(1) Standard voxel grids can be easily processed by 3D convolution for learning-based 3D tasks~\cite{maturana2015voxnet,wu20153d,choy20163d,wu2016learning}.
However, restricted by its cubic space complexity, voxel representation can not scale to a high resolution, usually limited to $64^3$. 
Even with efficient data structures like octrees or multi-scale representations \cite{wang2017cnn,hane2017hierarchical,riegler2017octnetfusion,9294054}, 
these representations still face limitations in quality.
(2)
Point clouds, extracted from shape surfaces, offer an alternative 3D representation~\cite{fan2017point,wu2019pointconv,qi2017pointnet,Xie_2021_CVPR}. They are memory-efficient and not restricted by resolution, unlike voxel representations. However, point clouds cannot represent topological relations in 3D space, and recovering shape surfaces from point clouds is nontrivial.
(3)
Recently, implicit  representations have gained attention for their simplicity in representing 3D shapes \cite{park2019deepsdf,mescheder2019occupancy,michalkiewicz2019implicit,sitzmann2020metasdf}. They work by predicting the signed distance \cite{park2019deepsdf,sitzmann2020metasdf} or occupancy label \cite{mescheder2019occupancy,peng2020convolutional} of a given point, allowing for easy surface recovery using Marching cubes \cite{lorensen1987marching} methods. Subsequent works \cite{mescheder2019occupancy,park2019deepsdf,chen2019learning,peng2020convolutional,jiang2020local} focus on designing implicit function representation with global or local shape priors.
To sum up, various 3D representations have led to diverse shape generative models, with notable examples including 3DGAN \cite{wu2016learning}, PC-GAN \cite{achlioptas2018learning}, IM-GAN \cite{chen2019learning},  and GBIF \cite{ibing20213d}. 
However, most current generative methods are task-specific. And it is difficult to be directly applied to  different generative tasks (\textit{e.g.}, shape completion).
Essentially, they rely on GANs for the generation step, suffering from known drawbacks such as mode collapse and training instability. In contrast, we propose an improved auto-regressive model for 3D shape generation that can synthesize high-quality and diverse shapes while being easily generalized to other multi-modal conditions.

\noindent \textbf{3D Auto-regressive Models}.
Auto-regressive models are probabilistic generative approaches that have tractable probability density. Using the probability chain rule, the likelihood of a given sample especially for high dimensional data, can be factorized into a series of products of conditional probability. In contrast, GANs do not have such a tractable probability density.
Recently,  AR models achieve 
remarkable progress in 2D image generation  \cite{esser2021taming,van2016conditional,razavi2019generating}, to a less extent 3D tasks~\cite{sun2020pointgrow, cheng2022autoregressive, nash2020polygen}.
Most of 3D AR models are struggling with generating high-quality shapes due to the challenges of representation learning with more points or faces.
Particularly, we notice two recent great works~\cite{yan2022shapeformer,mittal2022autosdf} that share similar insights into utilizing AR models for 3D tasks. Yan et al.~\cite{yan2022shapeformer} introduces a sparse representation to only quantize non-empty grids in 3D space, but still follows a monotonic row-major order. Mittal et al.~\cite{mittal2022autosdf} presents a non-sequential design to break the orders, but performs on all volumetric grids.
However, both of them address only one of the above limitations, and burden the structure design and training of transformer. 
In contrast, our Argus3D applies discrete representation learning in a latent vector instead of volumetric grids. Such a representation offers plenty of advantages, including shorter lengths of discrete codes, tractable orders from auto-regression, fast convergence, and also preserving essential 3D information. Moreover, benefiting from the simplicity of transformers, we can freely switch from unconditional generation to conditional generation by concatenating various conditions.

\noindent \textbf{3D Diffusion Models}.
Recently, Denoising Diffusion Probabilistic Models (DDPMs)~\cite{ldm,ddpm,ddim,dhariwal2021diffusion} have gained considerable attention for their stability and diverse image generation capabilities. Significant progress has also been made in 3D generation through point clouds based methods or 2D lift 3D~\cite{lion, cheng2023sdfusion, rodin, luo2021diffusion, zhou20213d,meshdiffusion,Diffusion-Based,zhang20233dshape2vecset,ye2023consistent,liu2023one}. DDPMs learn the data distribution from a Gaussian distribution through a gradual denoising process, often based on the U-Net~\cite{unet} architecture. However, this approach limits the resolution and extends the training cycles, thus hindering its performance in 3D generation.
In contrast, transformer-based models offer a high degree of scalability, which has been shown to enhance performance across various downstream tasks by scaling up the model size\cite{wei2022emergent,gpt4,zhai2022scaling,dehghani2023scaling}. In this paper, we posit that this scalability can also be advantageous in the domain of 3D shape generation, enabling more effective and efficient modeling of complex shapes.

\section{
Objaverse-Mix: Scaling Up Dataset with Assembled 3D Shapes
\label{sec:data}}

We draw inspiration from the significant progress made in the realm of 2D vision, where the efficacy of large models supported by extensive training data has been convincingly demonstrated
~\cite{dosovitskiy2021image, liu2021swin, fang2022eva}. However, it is nontrivial to build a large-scale 3D dataset.
On  one hand, for the datasets that are commonly used today, ShapeNet~\cite{shapenet} or ModelNet40~\cite{modelnet} only has tens of thousands 3D shapes, whose scale is too small to train a very large model, facing the risk of overfitting.
On the other hand, the process of labeling 3D objects involves considerably higher costs and complexities, making it one of the inherent challenges associated with constructing large-scale 3D datasets. Recently, Deitke \textit{et al.}~\cite{objaverse} present a massive dataset Objaverse with more than 800K 3D models, which are collected from an online 3D marketplace. Though Objaverse improves 3D repositories in terms of scale, it is still very  noisy and does not have 3D annotations, preventing the model from directly learning 3D knowledge from the data. The recent expansion of the Objaverse dataset has culminated in Objaverse-XL~\cite{objaversexl}, now featuring 10 million shapes. The sheer volume of data scale is very impressive, while we notice that  it still does not have very detailed 3D annotations and  multi-modal properties. Additionally, we also notice that some sourcing data from other sites such as GitHub and Thingiverse are still noisy to some extent.

In this paper, we find a simple and different way to build a large-scale 3D dataset, providing sufficient data for Argus3D model training. Concretely, we leverage an ensemble of five publicly-available 3D repositories, resulting in an impressive collection of approximately 900K 3D shapes.
This not only allows us to efficiently scale the data, but also requires almost no set-up cost. We thus name the dataset \textit{Objaverse-Mix}, as we acknowledge that a large portion of data is inherited and cleaned from Objaverse dataset.
We provide the details of each 3D repository below, as shown in Fig.~\ref{fig:datateaser}a. 

\noindent \textbf{ModelNet40}\cite{modelnet}: This dataset stands as a valuable resource, containing an impressive collection of over 12,300 CAD models spanning 40 object categories. For our research objectives, we strategically select 9,843 models from this dataset to serve as the foundation for training our model.

\noindent \textbf{ShapeNet}\cite{shapenet}: Recognized for its comprehensiveness and scope, ShapeNet provides an extensive and diverse collection of 3D models. Encompassing 55 common object categories and boasting over 51,300 unique 3D models, this dataset offers rich and varied representations that greatly enhance the training process of our model. The selection of ShapeNet bolsters the ability of our model to capture a wide range of object variations.

\noindent \textbf{Pix3D}\cite{pix3d}: With its large-scale nature, Pix3D emerges as a valuable resource for our research. Comprising 10,069 images and 395 shapes, this dataset exhibits significant variations, allowing our model to learn from a diverse range of object appearances and characteristics. By incorporating Pix3D into our training dataset, we equip our model with the capability to generate visually impressive and realistic 3D shapes.

\noindent \textbf{3D-Future}\cite{3dfuture}: As a specialized furniture dataset, 3D-Future provides a unique and detailed collection of 16,563 distinct 3D instances. By including this dataset in our training pipeline, we ensure that our model gains insights into the intricacies and nuances of furniture designs. This enables our model to generate high-quality and realistic 3D shapes in the furniture domain.

\noindent \textbf{Objaverse}\cite{objaverse}: A recent addition to the realm of 3D shape datasets, Objaverse captivates us with its vastness and richness. Sourced from Sketchfab, this immensely large dataset comprises over 800,000(and growing) 3D models, offering a wide array of object categories and variations. This extensive dataset provides our model with a wealth of diverse and representative examples for learning and generating high-quality 3D shapes.

Furthermore, to ensure the quality of 3D objects, we systematically utilize a series of pre-processing strategies. 
We first follow previous efforts \cite{choy20163d,occupancy} to normalize and re-scale all shapes to a uniform standard. Then, we employ the established techniques \cite{meshfusion} to render each 3D object, create depth maps, generate watertight meshes, and sample points from these meshes. This step helps us effectively to prune shapes that do not have watertight meshes. Next, we manually filter out noisy samples (\textit{e.g.}, weird or supernatural objects) or difficult examples (\textit{e.g.}, 3D scenes with lawns), which are empirically found to be detrimental to model convergence. To facilitate the learning of multi-modal generation, we also utilize the off-the-shelf tools \cite{blender,li2023blip} to render multi-view images and produce text descriptions. For each 3D object, we use Blender to render 12 views images in 512×512 resolution. Contrary to the approach employed by~\cite{zero123}, which utilizes random views for rendering, we opted for fixed perspectives to render images. This includes the front, side, top, back, and bottom views, along with three equidistant points at polar angles of 45° and 135°. This method of rendering helps avoid low-quality images and ensures consistency, particularly with the front-view renderings. The front view is particularly useful as it generally contains the most information about a shape's appearance. In addition, using fixed perspectives, especially the front view, allows us to generate higher-quality text captions. Random views can sometimes lead to inconsistent captions due to varying information presented in each view. To generate text captions for each 3D shape, we employ BLIP2~\cite{li2022blip}, using the images rendered from the front view. This approach ensures that our text captions are coherent and accurately reflect the content of the 3D shapes.
As a result, we construct a comprehensive mixed dataset Objaverse-Mix, as illustrated in Fig.~\ref{fig:datateaser}, including 900K objects with multiple representations of meshes, point clouds and voxels, and various labels of occupancy, rendered images and captions. 
To facilitate data assembly and pre-processing, we leverage four machines, each featuring a 64-core CPU and 8 A100 GPUs, for more than four weeks. The process necessitates nearly 100TB of storage owing to its intricate nature.

\section{Argus3D: Improved Auto-regressive Models for 3D Shape Generation \label{sec:method}}

\begin{figure*} 
\begin{centering}
\includegraphics[scale=0.47]{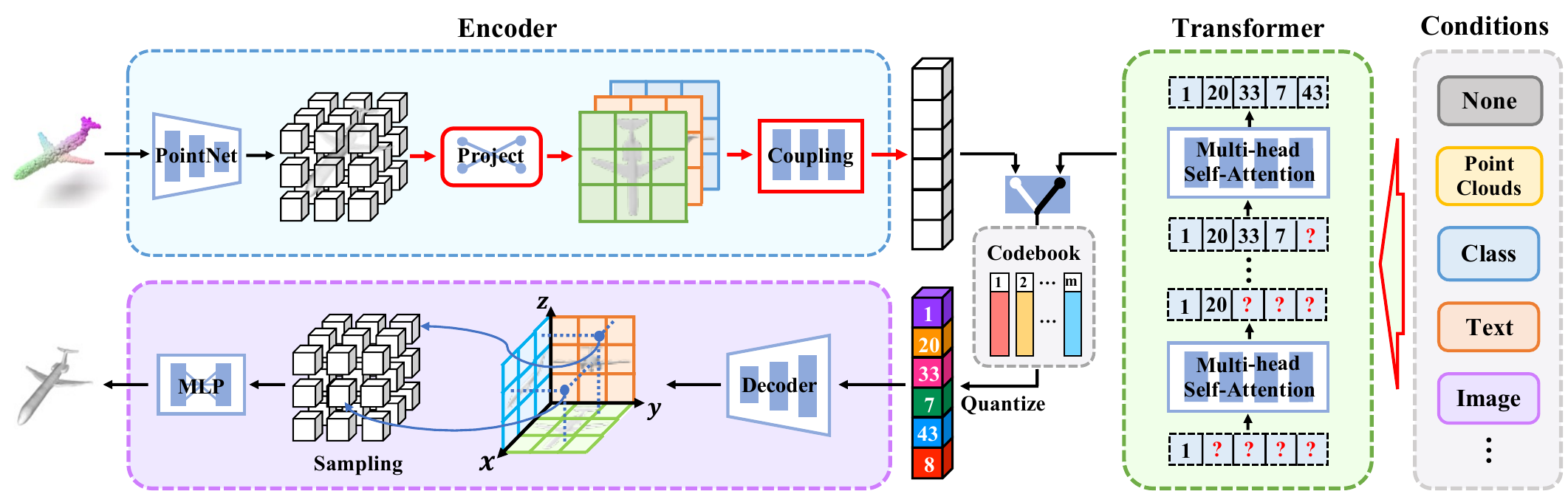}
 \vspace{-0.01in}
\caption{Overview of our Argus3D. Given an arbitrary 3D shape, we first project encoded volumetric grids into the three axis-aligned planes, and then use a coupling network to further project them into a latent vector. Vector quantization is thus performed on it for discrete representation. Taking advantage of such a compact representation with tractable orders, vanilla transformers are adopted to auto-repressively learn shape distributions. Furthermore, we can freely switch from unconditional generation to conditional generation by concatenating various conditions, such as point clouds, categories and images.} \label{fig:framework}
\vspace{-0.2in}
\end{centering}
\end{figure*}

Figure~\ref{fig:framework} illustrates the schematic of our proposed framework for 3D shape generation with the necessary symbol definitions and necessary preliminaries in Sec.~\ref{subsec:preliminary}.
From the perspective of capacity, it consists of a two-stage training procedure. 
We first represent the input as a composition of learned discrete codes as in Sec.~\ref{sec:stage1}; then utilize a transformer model to learn their interrelations in Sec.~\ref{sec:stage2}. 
From the perspective of scalability, Section~\ref{sec:large_model} takes advantage of our improved representation, and thus scales up both transformer layers and data for stronger generation ability.

\subsection{Preliminary \label{subsec:preliminary}}

\begin{figure}[t]
\begin{centering}
\includegraphics[scale=0.42]{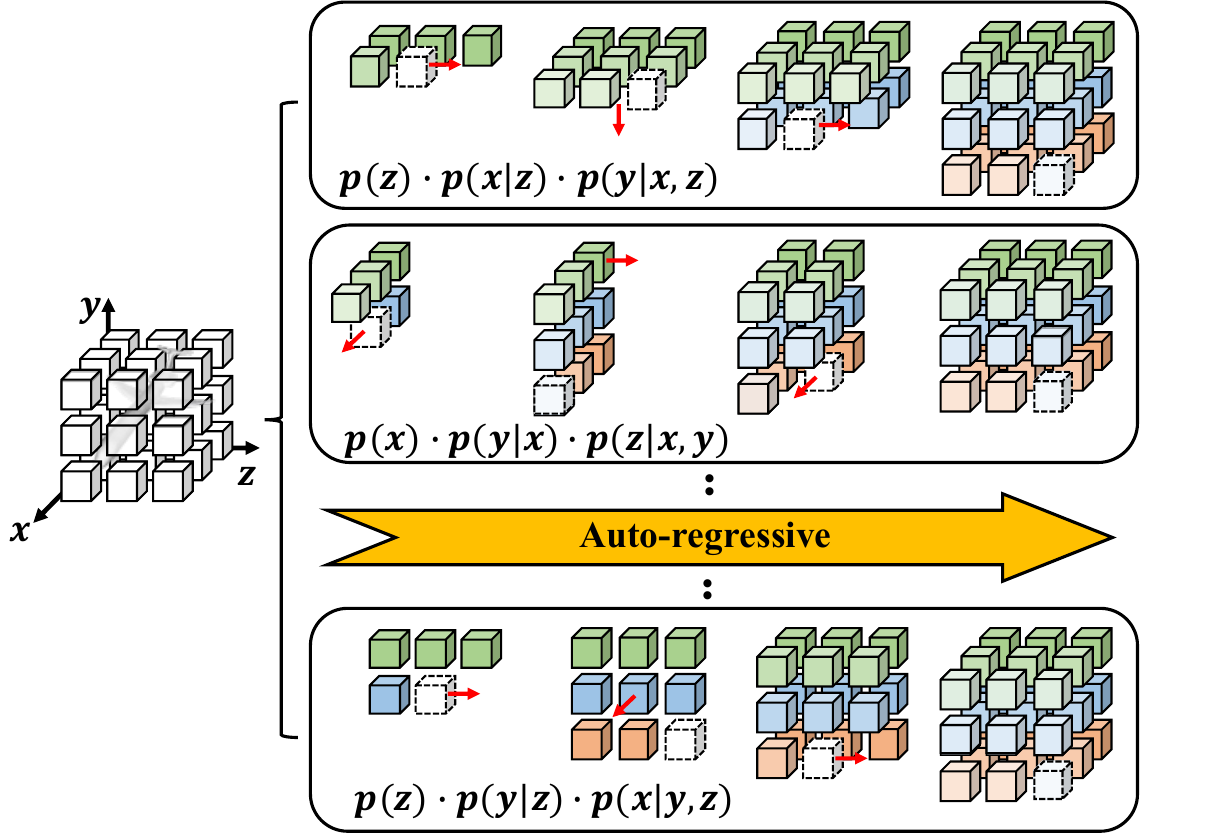}  
\tabularnewline
\vspace{-0.1in}
\caption{Illustration of auto-regressive generation for grid-based representation. Here, we show three different flattening orders as examples. Best viewed in color.
}
\label{fig:flatten_order}
\vspace{-0.2in}
\end{centering}
\end{figure}

\noindent \textbf{Ambiguity.} We highlight the problem of ambiguity existed in our task. 
Formally, \textit{`ambiguity' appears in the order of a series of conditional probabilities, which affects the difficulty of likelihood learning, leading to approximation error of the joint distribution.} 
Critically, auto-regressive models require sequential outputs, auto-regressively predicting the next code conditioned on all previous ones. Thereby, the order of the flattened sequence determines the order of conditional probabilities.
Although some methods such as position embedding~\cite{vaswani2017attention} can be aware of positions of codes, it cannot eliminate approximation error caused by the condition order. 
Notably, such the `ambiguity' phenomenon is also discussed in \cite{esser2021taming} and visualized by Fig.~\textcolor{red}{34} in \cite{esser2021taming}, where the loss curves in Fig.~\textcolor{red}{34} highlight the differences in difficulty for likelihood learning across various orders.
Figure~\ref{fig:flatten_order} illustrates how the flattening order affects the way of auto-regressive generation.
For grid-based representation, it is ambiguous if the flattening order along axes is $y$-$x$-$z$, $x$-$z$-$y$ or other combinations.

\noindent \textbf{Discrete Representation.}
Given input point clouds $P \in \mathbb{R}^{n\times3}$ where $n$ means the number of points, an encoder is adopted to extract features for each point cloud and then perform voxelization to get features of regular volumetric grids $f^{v} \in \mathbb{R}^{r\times r \times r\times c}$, where $r$ denote the resolution of voxels and $c$ is the feature dimension. To learn discrete representation for each 3D shape, a codebook $\mathbf{q}\in \mathbb{R}^{m\times c}$ is thus introduced whose entry is a learned code describing a particular type of local-part shape in a grid. Formally, for each grid $\left\{ f_{(h,l,w)}^{v} \right\}_{h,l,w=1}^{r}$, vector quantization $\mathcal{Q}\left(\cdot\right)$ is performed by replacing it with the closest entry in codebooks \cite{esser2021taming},
\begin{equation}
   \mathbf{z}^{v} = \mathcal{Q}\left(f^{v}\right) := \arg\min_{\mathbf{e}_{i}\in \mathbf{q}} || f^{v}_{\left(h,l,w\right)} - \mathbf{e}_{i} ||
\label{eq:quantize}
\end{equation}

\noindent where $\mathbf{e}_{i}\in \mathbf{q}$ represents the $i$-th entry in the codebook. 
Thus, learning the correlations between entries in the second stage can explore the underlying priors for shape generation. Nonetheless, auto-regressive generation \cite{esser2021taming,yan2022shapeformer} requires sequential outputs, facing two limitations. \textit{First, the resolution of $\mathbf{z}^{v}$ matters the quality of synthesized shapes}. Too small $r$ may lack the capacity to represent intricate and detailed geometries. A large value of $r$ can learn a specific code for each local gird. This will inevitably increase the computational complexity since the number of required codes explodes as $r$ grows. \textit{Second, the order of $\mathbf{z}^{v}$ affects the generation quality}. Each grid is highly coupled with neighbors. Simply flattening e.g. along x-y-z axes may cause the  problem of ambiguity, leading to sub-optimal generation quality.

\subsection{Improved Discrete Representation Learning \label{sec:stage1}}

One possible solution to solve the first limitation 
is applying vector quantization in spatial grids instead of volumetric grids inspired by \cite{chan2021efficient}. Specifically, after obtaining point cloud features $f^{p} \in \mathbb{R}^{n \times c}$, we first project points onto three axis-aligned orthogonal planes. Features of points falling into the same plane grid are aggregated via summation, resulting in three feature maps for the three planes $\left\{ f^{xy}, f^{yz}, f^{xz} \right\} \in \mathbb{R}^{l\times w\times c}$. 
Next, the vector quantization is applied to the three planes separately.
The primary advantage of tri-planar representation is efficient and compact. It dramatically reduces the number of grids from $\mathcal{O}(r^3)$ to $\mathcal{O}(r^2)$
while preserving essential 3D information. However, it still suffers from
the order ambiguity.  
Even worse, it involves the flattening order of three planes and the order of entries of each plane.

To this end, we further introduce a projection by learning a higher latent space for features of the three planes. This is simply achieved by first concatenating three planes with arbitrary order and then feeding them into a coupling network. Finally, the output is flattened as a projected latent vector, formulated as,

\begin{equation}
f = \tau\left(\mathcal{G}\left(\left[f^{xy};~f^{yz};~f^{xz} \right];~\theta\right)\right) \in \mathbb{R}^{m \times d}
\label{eq:laten_vector}
\end{equation}

\noindent where $\left[\cdot;\cdot\right]$ means the concatenation operation; $\mathcal{G}\left(\cdot;\theta\right)$ is a series of convolution layers with parameters $\theta$; $\tau\left(\cdot\right)$ is the flatten operation with row-major order;
$m$ and $d$ indicate the length of the latent vector and feature dimension. By applying discrete representation learning in the latent vector, we can describe each 3D shape with $\mathbf{z} = \mathcal{Q}\left(f\right)$, where $\mathcal{Q}\left(\cdot\right)$ is vector quantization in Eq.~\ref{eq:quantize}.

\noindent \textbf{Remark.} Different from existing works \cite{mittal2022autosdf,yan2022shapeformer} that rely on structure design and training strategies in the second stage to address the problem of ambiguous order, we tackle it by learning the coupling relationship of spatial grids in the first stage with the help of the second projection. By stacking convolution layers, we increase the receptive field of each element in the latent vector. Additionally, since the features of each spatial grid on the three planes are fused and highly encoded, each element does not have a definite position mapping in 3D space, which results in a tractable order for auto-regression. More in-depth discussions can be found in Sec.~\ref{sec:more_results}.

\noindent \textbf{Training Objective.} We optimize parameters with reconstruction loss. After getting discrete representation $\mathbf{z}$ which represents indices of entries in the codebook, we retrieve the corresponding codes with indices, denoted as $\mathbf{q}_{\left(\mathbf{z}\right)}$. Subsequently, a decoder with symmetric structures of the encoder is designed to decode $\mathbf{q}_{\left(\mathbf{z}\right)}$ back to features of the three planes.
Given sampling points $x\in \mathbb{R}^{3}$, we query their features by projecting them onto each of the three feature planes and performing bilinear interpolation. Features from three planes are accumulated and fed into an implicit function to predict their occupancy values. Finally, we apply binary cross-entropy loss between the predicted values $y_{o}$ and the ground-truth ones $\tilde{y_{o}}$,
\begin{equation}
   \mathcal{L}_{occ} = -\left( \tilde{y_{o}}\cdot\log\left(y_{o}\right) + \left(1-\tilde{y_{o}}\right)\cdot \log\left(1-y_{o}\right)   \right)
\end{equation}

To further train the codebook, we encourage pulling the distance between features before and after the vector quantization. Thus, the codebook loss is derived as,
\begin{equation}
   \mathcal{L}_{code} = \beta||\mathrm{sg}\left[ f\right] - \mathbf{q}_{\left(\mathbf{z}\right)} ||^{2}_{2} +  || f- \mathrm{sg}\left[ \mathbf{q}_{\left(\mathbf{z}\right)}\right]  ||^{2}_{2} 
\end{equation}

\noindent where $\mathrm{sg}\left[\cdot\right]$ denotes the stop-gradient operation \cite{van2017neural}. And we set $\beta=0.4$ by default. In sum, the overall loss for the first stage is $\mathcal{L}_{rec}=\mathcal{L}_{occ} + \mathcal{L}_{code}$.

\subsection{Learning Priors with Vanilla Transformers \label{sec:stage2}}

Benefiting from a compact composition and tractable order of discrete representation, models in the second stage can absorbedly learn the correlation between discrete codes,  effectively exploring priors of shape composition. We thus adopt a vanilla decoder-only transformer \cite{esser2021taming} without any specific-designed module.

\textit{For unconditional generation} and given discretized indices of latent vector $\mathbf{z} = \left\{\mathbf{z}_{1}, \mathbf{z}_{2}, \cdots, \mathbf{z}_{m} \right\}$, we feed them into a learnable embedding layer to retrieve features with discrete indices \footnote{We reuse the symbol of $\mathbf{z}$ after embedding for simplicity}. 
Then, the transformer with multi-head self-attention mechanism predicts the next possible index by learning the distribution of previous indices, $p\left(\mathbf{z}_{i}~|~\mathbf{z}_{<i}\right)$. This gives the joint distribution of full representation  as,
\begin{equation}
p\left(\mathbf{z}\right) = \prod_{i=1}^{m}p\left(\mathbf{z}_{i}~|~\mathbf{z}_{<i} \right)
\label{eq:uncond}
\end{equation}

\textit{For conditional generation}, users often expect to control the generation process by providing additional conditions. Instead of designing complex modules or training strategies, we simply learn joint distribution given conditions $\mathbf{c}$ by prepending it to $\mathbf{z}$. Equation~\ref{eq:uncond} is thus  extended as,
\begin{equation}
p\left(\mathbf{z}\right) = \prod_{i=1}^{m}p\left(\mathbf{z}_{i}~|~\mathbf{c},~\mathbf{z}_{<i} \right)
\label{eq:cond}
\end{equation}

\noindent where $\mathbf{c}$ denotes a feature vector of given conditions. The simplicity of our model gives the flexibility to learn conditions of any form. Specifically, for 3D conditions such as point clouds, we use our discrete representation learning in Sec.~\ref{sec:stage1} to transform them into a vector. As for 1D/2D conditions such as classes and images, we either adopt pre-trained models or embedding layers to extract their features.

\noindent \textbf{Objective.}
To train  second stage, we minimize  negative log-likelihood of Eq.~\ref{eq:uncond} or \ref{eq:cond} as $\mathcal{L}_{nll} = \mathbb{E}_{x \sim p\left(x\right)}\left[-\log p\left(\mathbf{z}\right)\right]$, where $p\left(x\right)$ is the distribution of real data.

\noindent \textbf{Inference.} With both models trained on two stages, we use Eq.~\ref{eq:uncond} or \ref{eq:cond} to perform shape generation by progressively sampling the next index with top-\textit{k} sampling strategy, until all elements in $\mathbf{z}$ are completed. Then, we feed $\mathbf{q}_{\left(\mathbf{z}\right)}$ into the decoder of the first stage, and query probabilities of occupancy values for all sampled 3D positions (\textit{e.g., $128^3$}). The output shapes are extracted with Marching Cubes \cite{lorensen1987marching}.

\subsection{More Parameters for Better Capacity
\label{sec:large_model}}
Recently, large models have shown remarkable advancements in both natural language processing and computer vision, such as GPT~\cite{gpt3,gpt4}, PaLM~\cite{chowdhery2023palm,anil2023palm}, BLIP~\cite{li2022blip,li2023blip} and EVA~\cite{fang2023eva,sun2023eva}. They not only have the ability to handle complex multi-tasks, but more importantly, play a vital pre-training role for downstream tasks.
Motivated by these successful works, we thus expect to enlarge the scale of our model, so as to simultaneously improve the capacity of 3D shape generation. 

Specifically, we increase the number of transformer layers and feature dimensions in the second stage, but retain the model parameters of the first stage for simplicity.
Since we utilize a vanilla decoder-only transformer here, the scale-up development is more effective with less effort.
Following the protocol of GPT-3 \cite{gpt3}, we design three variations of our model with different scales, \textit{i.e.}, Argus3D-B(ase), Argus3D-L(arge) and Argus3D-H(uge),
as demonstrated in Tab.~\ref{tab:para}.
Specifically, our Argus3D-H model has 32 transformer decoder layers, with more than 3.6 billion parameters. It is one of  the largest models in the field of 3D shape generation. 
Experimental results in Sec.~\ref{sec:scale} demonstrate that our scale-up model can further enhance the generation quality of 3D shapes.
This significant achievement in scaling up models marks a pivotal milestone, pushing the boundaries of 3D shape generation towards unprecedented advancements.
In the following paragraphs, we refer Argus3D to the base model unless otherwise specified.

\begin{figure}[t]
\begin{centering}
\includegraphics[scale=0.4]{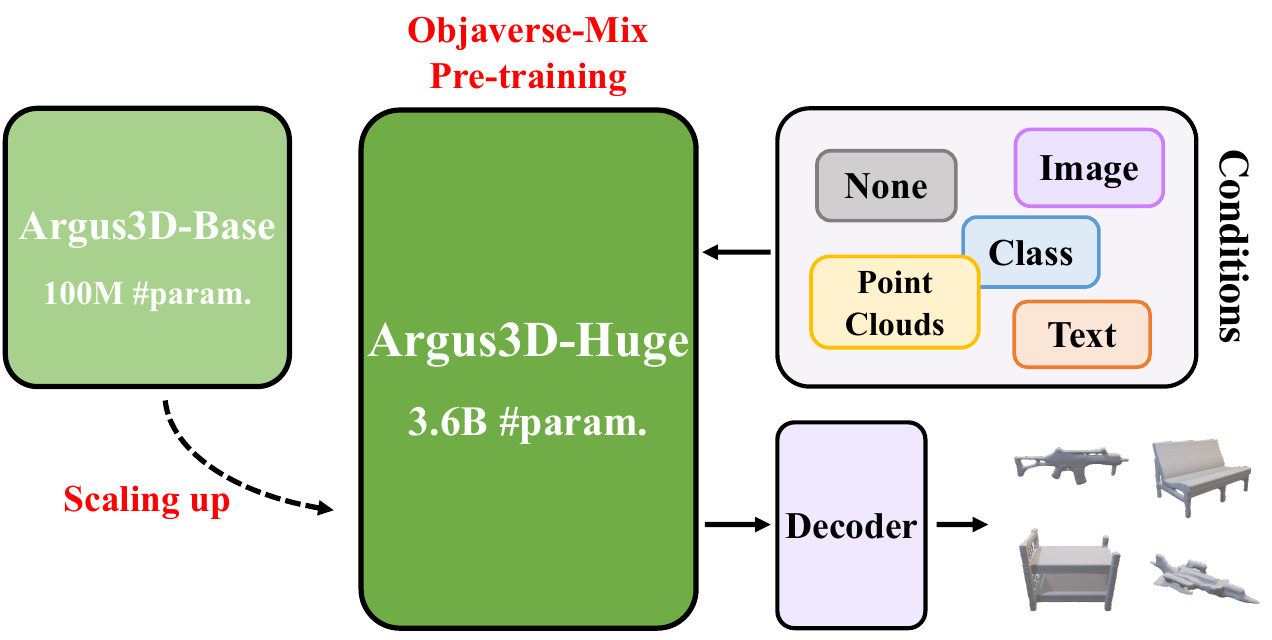}
\vspace{-0.1in}
\caption{Illustration of scaling up our models.
\label{fig:scale_up}}
\end{centering}
\end{figure}

\begin{table}[t]
\centering
\setlength{\tabcolsep}{3mm}{
\begin{tabular}{ccccc} 
\toprule
\multicolumn{1}{c}{\textsc{Size}} & \textsc{Dimensions} & \textsc{Layers} & \textsc{Heads} & \textsc{Params (M)} \tabularnewline 
\midrule
Base & 768 & 12 & 12 & 100   \tabularnewline 
Large & 2048 & 24 & 16 & 1239  \tabularnewline 
Huge & 3072 & 32 & 24 & 3670  \tabularnewline 
\bottomrule
\end{tabular}}
\vspace{-0.02in}
\caption{Three designs of our Argus3D with different scales.} 
\label{tab:para}
\vspace{-0.1in}
\end{table}

\section{Experiments \label{sec:experiments}}
This section starts with evaluating the capacity of our Argus3D. For a fair comparison, we show extensive studies on five generation tasks, demonstrating the powerful and flexible ability of Argus3D-B. (in Sec.~\ref{sec:shape_generation} $\sim$ \ref{sec:text_generation}). Next, we compare and discuss the efficacy of Argus3D at different scales  (in Sec.~\ref{sec:scale}). Lastly, we provide in-depth studies to evaluate the efficacy of our modules and show generalization to real-world data and zero-shot generation (in Sec.~\ref{sec:more_results}). For all experiments, if necessary, we sample point clouds from output meshes with Poisson Disk Sampling, or reconstruct meshes with our auto-encoder from output points, which is better than Poisson Surface Reconstruction \cite{kazhdan2006poisson}. 
Please refer to the Appendix for more implementation details.

\subsection{Evaluation Metrics \label{sec:metric}}
\noindent \textbf{Common metrics.}
As a generator, the key to evaluating our proposed method is not only to measure the \textit{fidelity}, but also to focus on the \textit{diversity} of the synthesized shapes. Therefore, we adopt eight metrics for different generation tasks, including Coverage (COV) \cite{achlioptas2018learning}, Minimum Matching Distance (MMD) \cite{achlioptas2018learning}, Edge Count Difference (ECD) \cite{ibing20213d} and 1-Nearest Neighbor Accuracy (1-NNA) \cite{yang2019pointflow}, Total Mutual Difference (TMD) \cite{wu2020multimodal}, Unidirectional Hausdorff Distance (UHD) \cite{wu2020multimodal}, Fr$\acute{\text{e}}$chet Point Cloud Distance (FPD) \cite{shu20193d} and Accuracy (Acc.) \cite{sanghi2022clip}. 
In particular, we use the Light Field Descriptor (LFD) \cite{chen2003visual} as our primary similarity distance metric for COV, MMD and ECD, as suggested by \cite{chen2019learning}. Since both FPD and Acc. metrics require a classifier to calculate, we thus train a PointNet \footnote{\url{https://github.com/jtpils/TreeGAN/blob/master/evaluation/pointnet.py}} with 13 categories on ShapeNet datasets, which achieves the classification accuracy of 92\%.

For shape generation task, COV and MMD measure the diversity and fidelity of the generated shapes, respectively. Both suffer from some drawbacks \cite{yang2019pointflow}. FPD and Acc. measure the fidelity of the generated shapes from the viewpoint of feature space and probability, respectively. On the contrary, ECD and 1-NNA measure the distribution similarity of a synthesized shape set and a ground-truth shape set in terms of both diversity and quality. Therefore, ECD and 1-NNA are two more reliable and important metrics to quantify the shape generation performance.  For shape completion tasks, TMD is meant to the diversity of the generated shapes for a partial input shape, and UHD is proposed to evaluate the completion fidelity. Both metrics are specifically designed for the completion task \cite{wu2020multimodal}.

\begin{figure}
\begin{centering}
\includegraphics[scale=0.24]{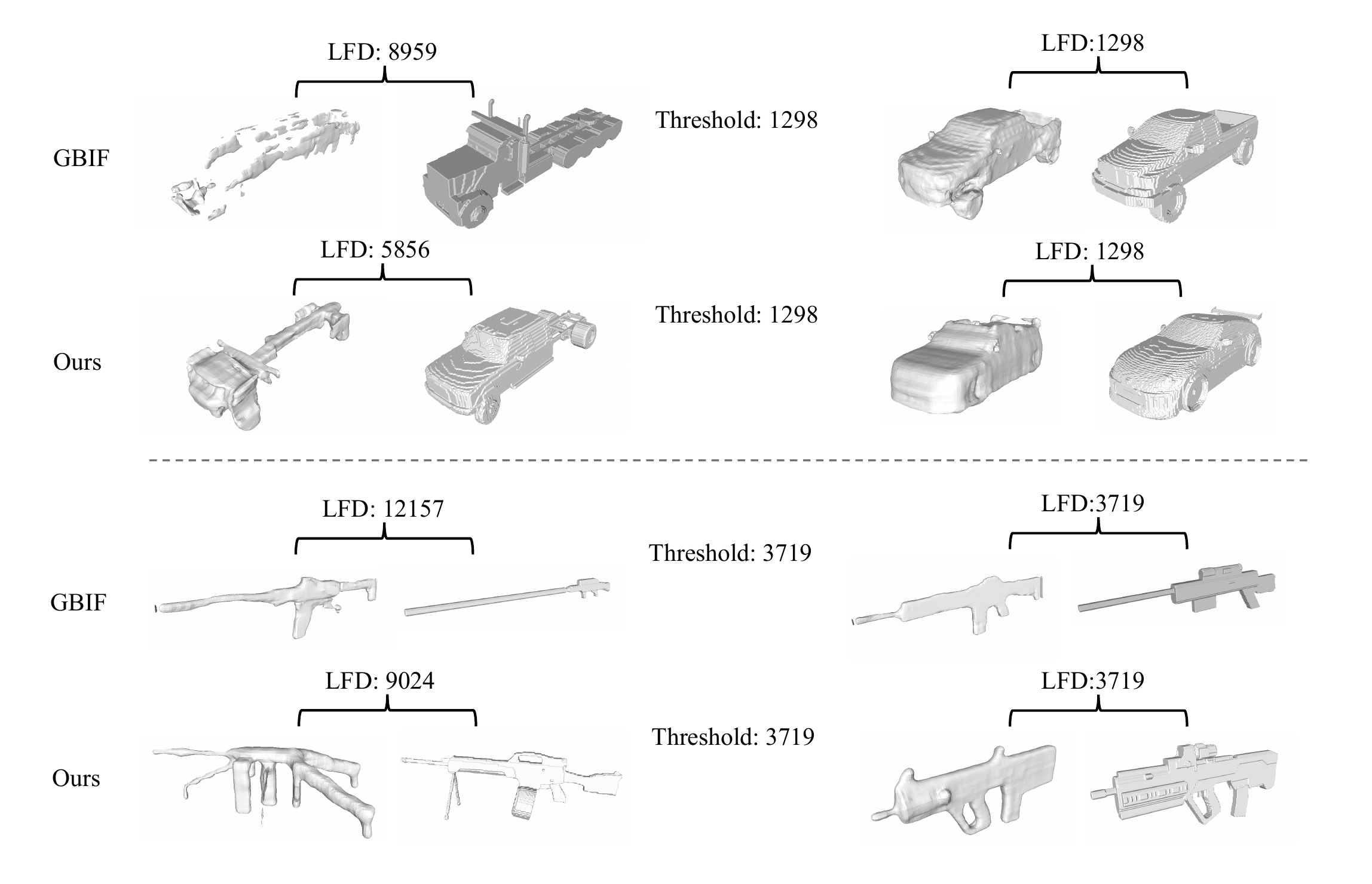}
\vspace{-0.1in}
\caption{We show the matched pairs with the largest MMD before and after using threshold shifting when computing the metric of \textit{Coverage}. For each pair, the left is the generated shape and the right is the closest ground-truth shape. Best viewed zoomed in.
\label{fig:t_cov}}
\vspace{-0.2in}
\end{centering}
\end{figure}

\noindent \textbf{\textit{Coverage} with threshold.}
As discussed above, Coverage \cite{achlioptas2018learning} measures the diversity of a generated shape set. However, it does not penalize outliers since a ground truth shape is still considered as covered even if the distance to the closest generated shape is large \cite{yang2019pointflow}. To rule out the false positive coverage, we introduce Coverage with threshold (CovT), to count as a match between a generation and ground truth shape only if LFD \cite{chen2003visual} between them is smaller than a threshold \textit{t}. In practice, \textit{t} could vary across different semantic categories based on the scale and complexity of the shape, and we empirically use MMD \cite{achlioptas2018learning} as the threshold. In this paper, we set \textit{t} as the mean MMD of all competitors.

To evaluate the effectiveness of the improved COV in identifying correct matches, we visualize the matched pairs with the largest MMD before and after using threshold filtering. As shown on the left of Fig.~\ref{fig:t_cov}, when there is no threshold constraining the quality of the generated shape, outliers (\textit{e.g.}, a messy shape) could match any possible ground-truth shape, which is clearly unreasonable. On the contrary, when the threshold is applied for filtering, as illustrated on the right of Fig.~\ref{fig:t_cov}, the generated shape has certain similarity in texture or parts with the matched ground-truth shape, even if they have the maximum shape distance. It strongly demonstrates the validity and reliability of our improvement in measuring the diversity of generations.

\subsection{Unconditional Shape Generation \label{sec:shape_generation}}

\noindent \textbf{Data.} We consider ShapeNet \cite{chang2015shapenet} as our main dataset for generation, following previous literature \cite{ibing20213d,chen2019learning,sun2020pointgrow}. We use the same training split and evaluation setup from \cite{ibing20213d} for fair comparability. Five categories of car, chair, plane, rifle and table are used for testing. As ground truth, we extract mesh from voxelized models with $256^3$ resolution in \cite{hane2017hierarchical}.

\noindent \textbf{Baselines.} 
We compare Argus3D with five state-of-the-art models, including GAN-based IM-GAN\cite{chen2019learning} and GBIF \cite{ibing20213d}, flow-based PointFlow \cite{yang2019pointflow}, score-based ShapeGF \cite{cai2020learning} and diffusion-based PVD \cite{zhou20213d}. 
We train these methods on the same data split with the official implementation.

\noindent \textbf{Metrics and Settings.} 
The size of the generated set is 5 times the size of the test set, the same as \cite{chen2019learning,ibing20213d}.
Coverage with threshold (CovT), Coverage (COV) \cite{achlioptas2018learning}, Minimum Matching Distance (MMD) \cite{achlioptas2018learning} and Edge Count Difference (ECD) \cite{ibing20213d} are adopted to evaluate the diversity, fidelity and overall quality of synthesized shapes. We also use 1-Nearest Neighbor Accuracy (1-NNA) \cite{yang2019pointflow} (with Chamfer Distance) to measure the distribution similarity. The number of sampled points is 2048.

\begin{table}
\centering
\setlength{\tabcolsep}{1mm}{
\footnotesize
\begin{tabular}{clcccccc}
\toprule
\multirow{2}{*}{\textsc{Metrics}} & \multicolumn{1}{c}{\multirow{2}{*}{\textsc{Methods}}} & \multicolumn{5}{c}{\textsc{Categories}} & \multirow{2}{*}{\textsc{Avg}}  \tabularnewline
\cmidrule{3-7}
 & & Plane & Car & Chair & Rifle & Table &  \tabularnewline 
\midrule
\multirow{6}{*}{ECD $\downarrow$} 
& IM-GAN \cite{chen2019learning} & \underline{923} & 3172 & 658 & 371 & 418 & 1108  \tabularnewline 
& GBIF \cite{ibing20213d} & 945 & \underline{2388} & \underline{354} & \underline{195} & 411 & \underline{858} \tabularnewline 
& PointFlow \cite{yang2019pointflow} & 2395 & 5318 & 426 & 2708 & 3559 & 2881 \tabularnewline
& ShapeGF \cite{cai2020learning} & 1200 & 2547 & 443 & 672 & \underline{114} & 915 \tabularnewline
& PVD \cite{zhou20213d} & 6661 & 7404 & 1265 & 3443 & 745 & 3904 \tabularnewline
& \textit{Ours} & \textbf{236} & \textbf{842} & \textbf{27} & \textbf{65} & \textbf{31} & \textbf{240} \tabularnewline  
\midrule

\multirow{6}{*}{1-NNA $\downarrow$} 
& IM-GAN \cite{chen2019learning} & 78.18 & 89.39 & 65.83 & 69.38 & 65.31 & 73.62 \tabularnewline
& GBIF \cite{ibing20213d}  & 80.22 & 87.19 & 63.95 & 66.98 & 60.96 & 71.86 \tabularnewline
& PointFlow \cite{yang2019pointflow} & \underline{73.61} & 74.75 & 70.18 & 64.77 & 74.81 & 71.62 \tabularnewline
& ShapeGF \cite{cai2020learning} & 74.72 & \underline{62.81} & \underline{59.15} & \underline{60.65} & \underline{55.58} & \underline{62.58} \tabularnewline
& PVD \cite{zhou20213d} & 81.09 & \textbf{57.37} & 62.36 & 77.32 & 74.31 & 70.49 \tabularnewline
& \textit{Ours} & \textbf{59.95} & 76.58 & \textbf{57.31} & \textbf{57.28} & \textbf{54.76} & \textbf{61.17} \tabularnewline
\midrule

\multirow{6}{*}{COV $\uparrow$} 
& IM-GAN \cite{chen2019learning} & 77.01 & 65.37 & 76.38 & 73.21 & \underline{85.71} & 75.53 \tabularnewline
& GBIF \cite{ibing20213d} & \textbf{80.96} & \textbf{78.85} & \textbf{80.95} & \textbf{77.00} & 85.13 & \textbf{80.57} \tabularnewline
& PointFlow \cite{yang2019pointflow} & 65.64 & 64.97 & 57.49 & 48.52 & 71.95 & 61.71 \tabularnewline
& ShapeGF \cite{cai2020learning} & 76.64 & 71.85 & 79.41 & 70.67 & \textbf{87.54} & 77.22 \tabularnewline
& PVD \cite{zhou20213d} & 58.09 & 58.64 & 68.93 & 56.12 & 76.84 & 63.72 \tabularnewline
& \textit{Ours} & \underline{79.11} & \underline{73.25} & \underline{80.81} & \underline{74.26} & 84.01 & \underline{78.29} \tabularnewline
\midrule

\multirow{6}{*}{CovT $\uparrow$} 
& IM-GAN \cite{chen2019learning} & \underline{41.03} & 50.63 & \underline{45.68} & \underline{51.68} & 46.50 & 47.10 \tabularnewline
& GBIF \cite{ibing20213d} & 32.38 & 52.76 & 39.77 & 50.00 & 43.68 & 43.72   \tabularnewline
& PointFlow \cite{yang2019pointflow} & 35.85 & 47.76 & 28.48 & 34.81 & 30.98 & 35.57 \tabularnewline
& ShapeGF \cite{cai2020learning} & 40.17 & \underline{53.63} & 43.69 & 51.05 & \textbf{48.50} & \underline{47.41} \tabularnewline
& PVD \cite{zhou20213d} & 12.11 & 43.36 & 38.82 & 33.33 & 43.68 & 34.26 \tabularnewline
& \textit{Ours} & \textbf{45.12} & \textbf{56.64} & \textbf{49.82} & \textbf{55.27} & \underline{48.03} & \textbf{50.98} \tabularnewline
\midrule

\multirow{6}{*}{MMD $\downarrow$} 
& IM-GAN \cite{chen2019learning} & \underline{3418} & \underline{1290} & 2881 & \underline{3691} & 2505 & \underline{2757} \tabularnewline
& GBIF \cite{ibing20213d} & 3754 & 1333 & 3015 & 3865 & 2584 & 2910 \tabularnewline
& PointFlow \cite{yang2019pointflow} & 3675 & 1393 & 3322 & 4038 & 2936 & 3072 \tabularnewline
& ShapeGF \cite{cai2020learning} & 3530 & 1307 & \underline{2880} & 3762 & \underline{2420} & 2780 \tabularnewline
& PVD \cite{zhou20213d} & 4376 & 1432 & 3064 & 4274 & 2623 & 3154 \tabularnewline
& \textit{Ours} & \textbf{3124} & \textbf{1213} & \textbf{2703} & \textbf{3628} & \textbf{2374} & \textbf{2608} \tabularnewline 
\bottomrule
\end{tabular}}
\vspace{-0.02in}
\caption{Results of unconditional generation. Models are trained for each category. The best and second results are highlighted in \textbf{bold} and \underline{underlined}.}
\label{tab:unconditional}
\vspace{-0.2in}
\end{table}

\begin{figure}
\begin{centering}
\includegraphics[scale=0.08]{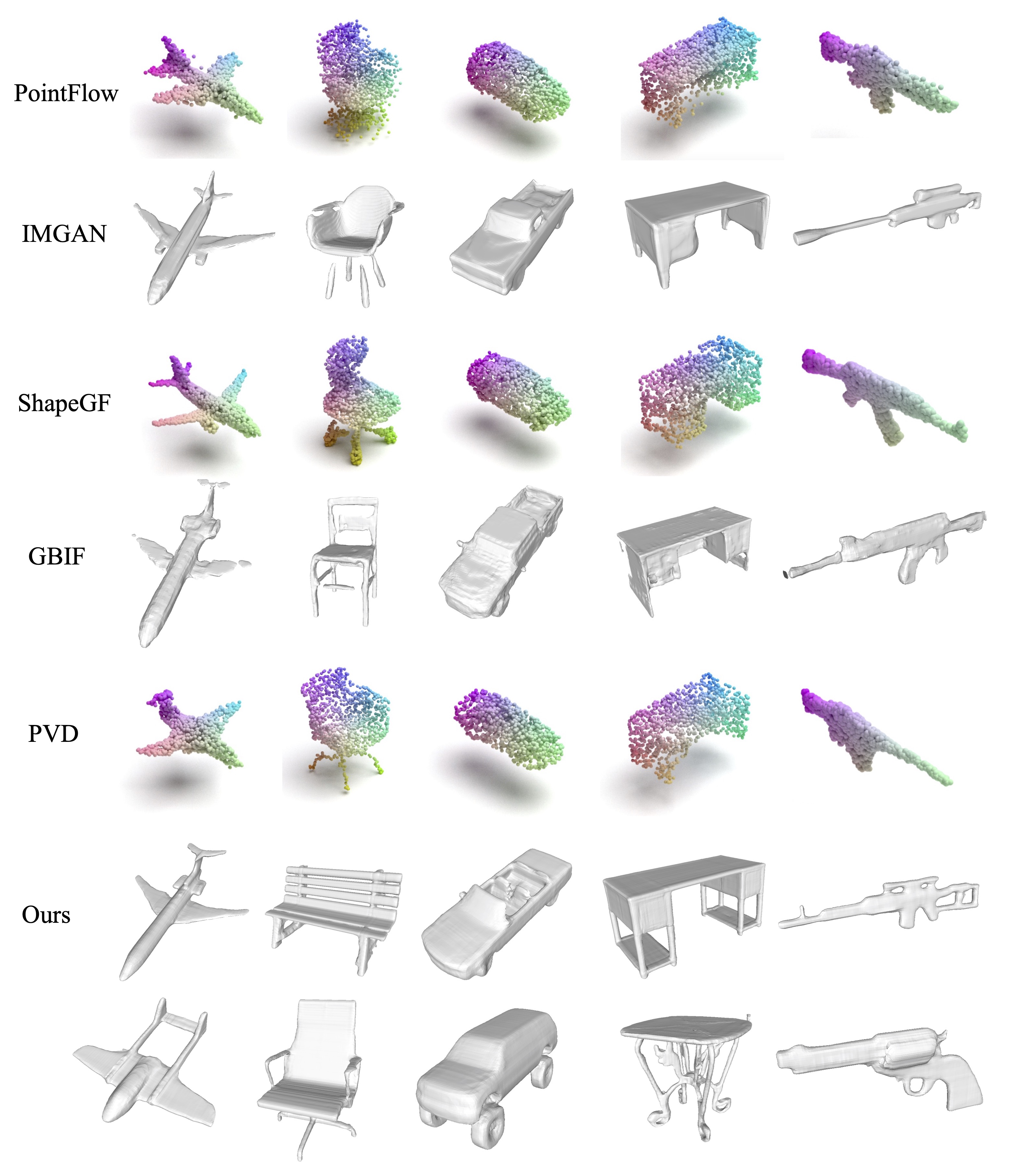}
\vspace{-0.1in}
\caption{Qualitative results of unconditional generation. \label{fig:unconditional} 
}
\vspace{-0.1in}
\end{centering}
\end{figure}

\noindent \textbf{Results Analysis.}
Results are reported in Tab.~\ref{tab:unconditional}. 
First, Argus3D achieves state-of-the-art performance with regard to ECD and 1-NNA. It significantly demonstrates the superiority of our model over synthesizing high-quality shapes. We notice that the result of Car is not good on the metric of 1-NNA. One possible reason is that Argus3D tries to generate meshes of tires and seats inside cars, which may not be very friendly to CD.
Second, our model has a clear advantage on both MMD and CovT metrics compared with all competitors, which separately indicates the outstanding fidelity and diversity of our generated shapes.
Third, though GBIF achieves relatively good results on COV, it gets worse results on CovT, suggesting that most of the matched samples come from false positive pairs. 
Argus3D, on the contrary, gets the second best performance on COV, but higher than GBIF on CovT by about 6 points, 
showing that our generated shapes enjoy the advantage of high-quality and fewer outliers. 
Lastly, we visualize shapes with multiple categories in Fig.~\ref{fig:unconditional}, further supporting the quantitative results and conclusions described above.

\subsection{Class-guide Generation \label{sec:class_generation}}
We evaluate the versatility of Argus3D on class-guide generation, which requires generating shapes given a category label. It is a basic conditional generation task. We use the same dataset and evaluation metrics as in Sec.~\ref{sec:shape_generation}.

\noindent \textbf{Baselines.} We choose two recently-published works as competitors due to similar motivation. One is a two-stage generative model GBIF \cite{ibing20213d}, and the other is AR method AutoSDF \cite{mittal2022autosdf}. We simply modify the mask-condition of \cite{ibing20213d} to class-condition, and additionally add the class token to transformers for \cite{mittal2022autosdf} which is the same as Argus3D.

\noindent \textbf{Results Analysis.} In Tab.~\ref{tab:class_cond}, Argus3D outperforms both competitors across 5 categories by a significant margin, achieving state-of-the-art results on all metrics. Particularly, it gets advantages on the metric of 1-NNA, strongly demonstrating the versatility
of Argus3D on class-guide generation. 
Qualitative results of 5 different categories are further illustrated in Fig.~\ref{fig:class_cond}. As observed, the generated quality of our method is clearly better than GBIF and AutoSDF, while preserving more diversity in types and shapes.

\begin{table}
\centering
\footnotesize 
\setlength{\tabcolsep}{1mm}{
\begin{tabular}{clcccccc}
\toprule
\multirow{2}{*}{\textsc{Metrics}} &  \multicolumn{1}{c}{\multirow{2}{*}{\textsc{Methods}}} & \multicolumn{5}{c}{\textsc{Categories}} & \multirow{2}{*}{\textsc{AVG}}  \tabularnewline
\cmidrule{3-7}
 & & Plane & Car & Chair & Rifle  & Table & \tabularnewline 
\midrule
\multirow{3}{*}{COV $\uparrow$} 
& GBIF \cite{ibing20213d} & 68.72 & 69.64 & 75.94 &   68.98 & 81.72 & 73.00 \tabularnewline
& AutoSDF \cite{mittal2022autosdf} & 70.46 & 52.77 & 63.25 &  48.10  & 72.19 & 61.35 \tabularnewline
& \textit{Ours} & \textbf{81.58} & \textbf{71.58} & \textbf{83.98} & \textbf{75.74} & \textbf{85.48} & \textbf{79.67} \tabularnewline

\midrule
\multirow{3}{*}{CovT $\uparrow$} 
& GBIF \cite{ibing20213d} & 24.10 & 38.63 & 32.69 & 35.44 & 37.80 & 33.73 \tabularnewline
& AutoSDF \cite{mittal2022autosdf} & 30.66 & 40.49 & 31.00 & 34.60 & 36.10 & 34.57 \tabularnewline
& \textit{Ours} & \textbf{56.49} & \textbf{52.70} & \textbf{45.09} & \textbf{52.74} & \textbf{49.32} & \textbf{51.27} \tabularnewline

\midrule
\multirow{3}{*}{MMD $\downarrow$} 
& GBIF \cite{ibing20213d} & 4736 & 1479 & 3220 & 4246 & 2763 & 3289 \tabularnewline
& AutoSDF \cite{mittal2022autosdf} & 3706 & 1456 & 3249 & 4115   & 2744 &  3054 \tabularnewline
& \textit{Ours} & \textbf{3195} & \textbf{1285} & \textbf{2871} & \textbf{3729} & \textbf{2430} & \textbf{2702} \tabularnewline

\midrule
\multirow{3}{*}{ECD $\downarrow$} 
& GBIF \cite{ibing20213d} & 1327 & 2752 & 1589 & 434 & 869 & 1394 \tabularnewline
& AutoSDF \cite{mittal2022autosdf} & 1619 & 4256 & 1038 &  1443  &  462 & 1764 \tabularnewline
& \textit{Ours} & \textbf{571} & \textbf{1889} & \textbf{419} & \textbf{196} & \textbf{285} & \textbf{672} \tabularnewline 

\midrule
\multirow{3}{*}{1-NNA $\downarrow$} 
& GBIF \cite{ibing20213d} & 91.47 & 92.43 & 75.61 & 83.12 & 70.19 & 82.56 \tabularnewline
& AutoSDF \cite{mittal2022autosdf} & 83.31 & 87.76 & 69.34 &  77.43  &  67.20 & 77.01 \tabularnewline
& \textit{Ours} & \textbf{66.81} & \textbf{83.39} & \textbf{64.83} & \textbf{57.28} & \textbf{59.55} & \textbf{66.37} \tabularnewline 

\bottomrule
\end{tabular}}
\vspace{-0.02in}
\caption{Results of class-guide generation. Models are trained on 13 categories of ShapeNet. Best results are highlighted in \textbf{bold}.}
\label{tab:class_cond}
\vspace{-0.1in}
\end{table}

\begin{figure}[t]
\begin{centering}
\includegraphics[width=8cm, height=8cm]{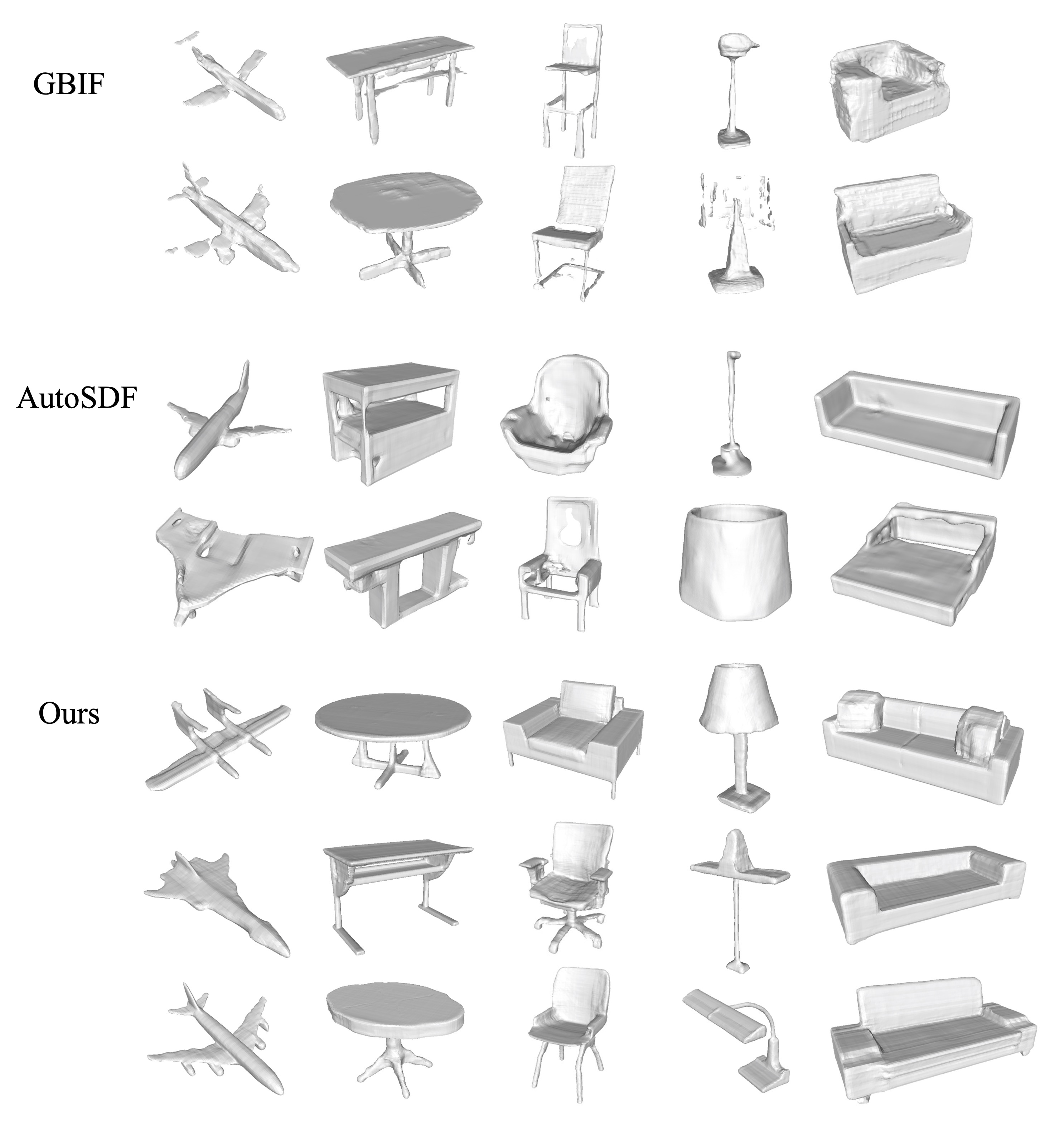}  
\tabularnewline
\vspace{-0.02in}
\caption{Qualitative results of class-guide generation.}
\label{fig:class_cond}
\vspace{-0.2in}
\end{centering}
\end{figure}

\begin{figure}[t]
 \begin{centering}
\includegraphics[scale=0.065]{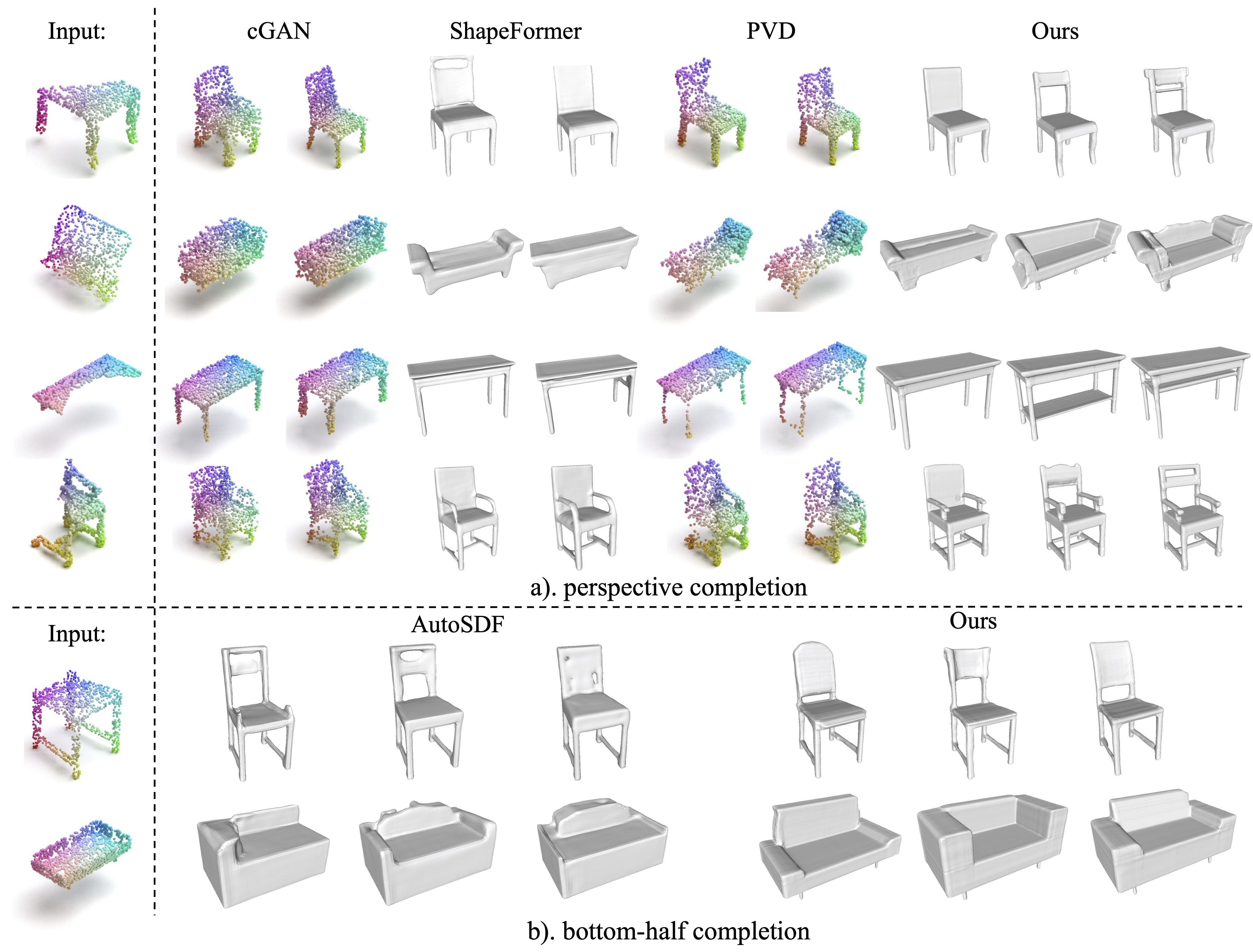}
\vspace{-0.02in}
\caption{Qualitative results of partial point completion. }
\label{fig:shape_completion}
\end{centering}
\end{figure}

\subsection{Multi-modal Partial Point Completion}

We further verify the ability of our model in conditional generation by giving partial point clouds. Here, we advocate the multi-modal completion, since there are many possibilities of the completed shape given the partial shape. It's the essence of generative models, where using your imagination while being faithful to the given partial conditions.

\noindent \textbf{Data.} 
We use ShapeNet dataset for testing. Two settings are considered here,
(1) perspective completion \cite{yu2021pointr}: randomly sampling a viewpoint and then removing the $25\% \sim 75\%$ furthest points from the viewpoint;
(2) bottom-half completion \cite{mittal2022autosdf}: removing all points from the top half of shapes.

\noindent \textbf{Baselines.} Four multi-modal completion models are chosen as baselines, including one generative adversarial model cGAN \cite{wu2020multimodal}, one diffusion model PVD \cite{zhou20213d}, two AR models ShapeFormer \cite{yan2022shapeformer} and AutoSDF \cite{mittal2022autosdf}.

\noindent \textbf{Metrics and Settings.} We complete $10$ samples for $100$ randomly selected shapes of three categories, \textit{i.e.}, chair, sofa and table.
Following \cite{wu2020multimodal}, we use Total Mutual Difference (TMD) to measure the diversity. Minimum Matching Distance \cite{achlioptas2018learning} (MMD) with Chamfer Distance and Unidirectional Hausdorff Distance (UHD) \cite{shu20193d} are adopted to measure the faithfulness of completed shapes.

\noindent \textbf{Results Analysis.} We first report perspective completion results in Tab.~\ref{tab:viewpoint_completion}. Argus3D beats all baselines and achieves state-of-the-art performance. Importantly, we outperform ShapeFormer on all classes and metrics, which also utilizes an AR model with transformers to learn shape distribution. On the other hand, we compare with AutoSDF in its bottom-half completion setting. Results from Tab.~\ref{tab:bottom_completion} illustrate that Argus3D outperforms AutoSDF in terms of TMD and MMD. It strongly suggests the flexibility and versatility of our proposed method.
Qualitative results in Fig.~\ref{fig:shape_completion} show the diversity and fidelity of our completed shapes.

\begin{table}
\centering
\footnotesize 
\setlength{\tabcolsep}{2.1mm}{
\begin{tabular}{clcccc}
\toprule
\multirow{2}{*}{\textsc{Metrics}} &  \multicolumn{1}{c}{\multirow{2}{*}{\textsc{Methods}}} & \multicolumn{3}{c}{\textsc{Categories}} & \multirow{2}{*}{\textsc{AVG}}  \tabularnewline
\cmidrule{3-5}
 & & Chair & Sofa & Table & \tabularnewline 
\midrule
& cGAN \cite{wu2020multimodal} & 1.708 & 0.687 & \textbf{1.707} & 1.367 \tabularnewline
TMD $\uparrow$ & PVD \cite{zhou20213d} & 1.098 & 0.811 & 0.839 & 0.916  \tabularnewline
($\times 10^{2}$) & ShapeFormer \cite{yan2022shapeformer} & 1.159 & 0.698 & 0.677 & 0.845 \tabularnewline
& \textit{Ours} & \textbf{2.042} & \textbf{1.221} & 1.538 & \textbf{1.600} \tabularnewline 

\midrule
& cGAN \cite{wu2020multimodal} & 7.836 & 7.047 & 9.406 & 8.096 \tabularnewline
UHD $\downarrow$ & PVD \cite{zhou20213d} & 10.79 & 13.88 & 11.38 & 12.02 \tabularnewline
($\times 10^{2}$) & ShapeFormer \cite{yan2022shapeformer} & 6.884 & 8.658 & 6.688 & 7.410 \tabularnewline
& \textit{Ours} & \textbf{6.439} & \textbf{6.447} & \textbf{5.948} & \textbf{6.278} \tabularnewline 

\midrule
& cGAN \cite{wu2020multimodal} & 1.665 & 1.813 & 1.596 & 1.691 \tabularnewline
MMD $\downarrow$ & PVD \cite{zhou20213d} & 2.352 & 2.041 & 2.174 & 2.189 \tabularnewline
($\times 10^{3}$) & ShapeFormer \cite{yan2022shapeformer} & 1.055 & 1.100 & 1.066 & 1.074 \tabularnewline
& \textit{Ours} & \textbf{0.961} & \textbf{0.819} & \textbf{0.828} & \textbf{0.869} \tabularnewline 
\bottomrule
\end{tabular}}
\vspace{-0.05in}
\caption{Results of multi-modal partial point completion. The missing parts vary according to random viewpoints.}
\label{tab:viewpoint_completion}
\end{table}

\begin{table}
\centering
\footnotesize
\setlength{\tabcolsep}{2.5mm}{
\begin{tabular}{clcccc} 
\toprule
\multirow{2}{*}{\textsc{Metrics}} & \multicolumn{1}{c}{\multirow{2}{*}{\textsc{Methods}}} & \multicolumn{3}{c}{\textsc{Categories}} & \multirow{2}{*}{\textsc{AVG}}  \tabularnewline  
\cmidrule{3-5}
&  & Chair & Sofa & Table &  \tabularnewline  
\midrule
TMD $\uparrow$ & AutoSDF  \cite{mittal2022autosdf} & 2.046  & 1.609 & 3.116 & 2.257  \tabularnewline 
\multicolumn{1}{c|}{($\times 10^2$)} & \textit{Ours} & \bf{3.682} & \bf{2.673} & \bf{10.30} & \bf{5.552}  \tabularnewline 
\midrule
UHD $\downarrow$ & AutoSDF \cite{mittal2022autosdf} & \bf{6.793} & 9.950 & \bf{8.122} & 8.289  \tabularnewline 
\multicolumn{1}{c|}{($\times 10^2$)}& \textit{Ours} & 6.996 & \bf{6.599} & 10.87 & \bf{8.155}  \tabularnewline 
\midrule
MMD $\downarrow$ & AutoSDF \cite{mittal2022autosdf}  & 1.501 & 1.154  & \bf{2.600} & \bf{1.751}  \tabularnewline 
\multicolumn{1}{c|}{($\times 10^3$)} & \textit{Ours} & \textbf{1.477} & \bf{1.127} & 2.906 & 1.837 \tabularnewline 
\bottomrule
\end{tabular}}
\vspace{-0.05in}
\caption{Quantitative results of multi-modal partial point completion. The missing parts are always the top half.}
\label{tab:bottom_completion}
\end{table}

\subsection{Image-guide Generation \label{sec:image_generation}}
Next, we show the flexibility that Argus3D can easily extend to image-guide generation, which is a more challenging task. The flexibility lies in that (1) it is implemented by the easiest way of feature concatenation; (2) the condition form of images is various, which could be 1-D feature vectors or patch tokens, or 2-D feature maps. For simplicity, we use a pre-trained CLIP model (\textit{i.e.}, ViT-B/32) to extract feature vectors of images as conditions. All experiments are conducted on ShapeNet with rendered images.

\noindent \textbf{Baselines.} CLIP-Forge \cite{sanghi2022clip} is a flow-based model, which is trained with pairs of images and shapes. We take it as our primary baseline for two reasons: (1) it is a generation model instead of a reconstruction model, and (2) it originally uses CLIP models to extract image features. We also compare with AutoSDF \cite{mittal2022autosdf}, which is an auto-regressive model for 3D shape generation. We follow its official codes to extract per-location conditionals from the input image and then guide the non-sequential auto-regressive modeling to generate shapes. 

\noindent \textbf{Metrics and Settings.} We evaluate models with the test split of 13 categories. For each category, we randomly sample 50 single-view images and then generate 5 shapes for evaluation. As a generation task, TMD is adopted to measure the diversity. We further use MMD with Chamfer Distance and Fr$\acute{\text{e}}$chet Point Cloud distance (FPD) \cite{shu20193d} to measure the fidelity compared with the ground truth.

\begin{figure}[t]
 \begin{centering}
\includegraphics[scale=0.38]{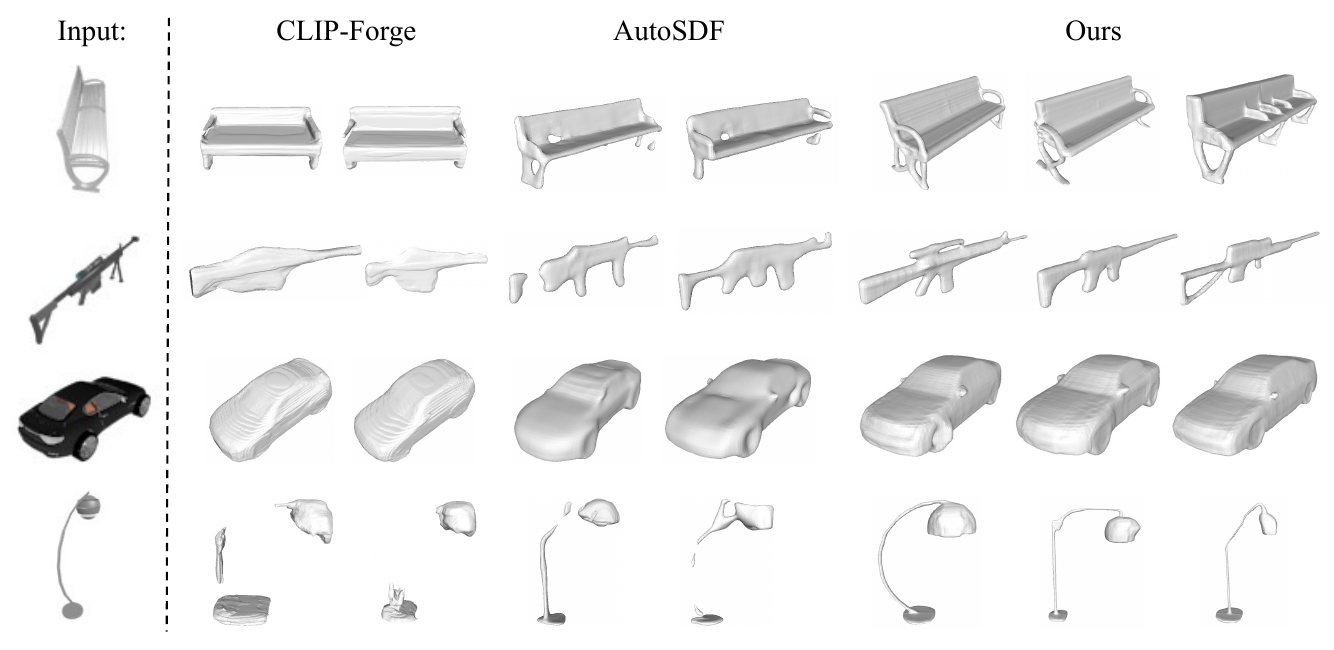}
\vspace{-0.02in}
\caption{Visualizations of image-guide shape generation.} \label{fig:image_generation}
\vspace{-0.05in}
\end{centering}
\end{figure}

\begin{table}[t]
\centering
\footnotesize
\setlength{\tabcolsep}{2.8mm}{
\begin{tabular}{lccc} 
\toprule 
\multicolumn{1}{c}{\textsc{Method}} & TMD ($\times 10^2 $) $\uparrow$ & MMD ($\times 10^3$) $\downarrow$ & FPD $\downarrow$ \tabularnewline 
\midrule
AutoSDF \cite{mittal2022autosdf} & 2.523 & \textbf{1.383} & 2.092 \tabularnewline 
Clip-Forge \cite{sanghi2022clip} & 2.858 & 1.926 & 8.094   \tabularnewline 
\midrule
\textit{Ours} (ViT32)  & 3.677 & 1.617 & 2.711  \tabularnewline
\textit{Ours} (ResNet) & \textbf{4.738} & 1.662 & 3.894 \tabularnewline
\textit{Ours} (CLIP)     & 4.274 & 1.590 & \textbf{1.680}  \tabularnewline 
\bottomrule
\end{tabular}}
\vspace{-0.08in}
\caption{Quantitative results of image-guide generation.} \label{tab:image_generation}
\vspace{-0.1in}
\end{table}

\noindent \textbf{Results Analysis.} Results are reported in Tab.~\ref{tab:image_generation}. Argus3D wins out in terms of both fidelity and diversity.
In particular, we achieve a great advantage on both TMD and FPD, demonstrating the effectiveness of Argus3D applied to image-guide generation. 
It is also successfully echoed by qualitative visualizations in Fig.~\ref{fig:image_generation}. 
Our samples are diverse and appear visually faithful to attributes of the object in images.
Most importantly, our approach can accept input in either 1D or 2D form, giving it more flexibility. 
To verify it, we conduct two baselines by using ViT32 (or ResNet) to extract patch embeddings (or image features) from the input image as conditions. The conditional generation can be achieved by simply prepending the condition to [SOS] token.
Results in Tab.~\ref{tab:image_generation} indicate that they are competitive with models using CLIP as the feature extractor, further suggesting the powerful versatility of our Argus3D in either generative ways or conditional forms.

\subsection{Text-guide Generation \label{sec:text_generation}}
Encouraged by promising results on image-guide generation, we also turn Argus3D to text-to-shape generation. The same pre-trained CLIP model is used to extract single embedding for text conditions. Note that we did it on purpose, not using word sequence embedding, but only using a single features vector in our model to show its efficacy and scalability to the simplest forms of conditions. We will discuss the length of text conditions later.

\noindent \textbf{Data.} One of important paired text-shape datasets is Text2Shape \cite{chen2018text2shape}, which provides language descriptions for two objects from ShapeNet, \textit{i.e.}, chair and table. Thereby, we consider it as our main dataset to perform text-guide shape generation.

\noindent \textbf{Baselines.} We compare our model with three state-of-the-art text-to-shape generation models. One is CLIP-Forge \cite{sanghi2022clip} under the supervised learning setting, using the same condition extractor as ours. One is ITG \cite{liu2022towards}, which adopts BERT to encode texts into sequence embeddings. And the other is AutoSDF \cite{mittal2022autosdf}, where we re-train it using its official codes.

\begin{table}[t]
\centering
\footnotesize
\begin{subtable}{1.\linewidth}
	\centering
	\footnotesize
	\setlength{\tabcolsep}{2.7mm}{
	\begin{tabular}{lccc} 
	\toprule 
	\multicolumn{1}{c}{\textsc{Method}} & TMD ($\times 10^1 $) $\uparrow$ & MMD ($\times 10^3$) $\downarrow$ & Acc $\uparrow$ \tabularnewline 
	\midrule
	ITG \cite{liu2022towards}  & N/A & 2.187  &  29.13  \tabularnewline 
        AutoSDF \cite{mittal2022autosdf} & 0.342 & 2.165 & 36.95 \tabularnewline 
	CLIP-Forge \cite{sanghi2022clip} & 0.400 & 2.136 &  53.68  \tabularnewline 
	\midrule
         \textit{Ours} (BERT) & \textbf{0.677} & 1.931 & \textbf{60.68} \tabularnewline
        \textit{Ours} (CLIP-seq.)  & 0.524 & \textbf{1.778} & 58.17 \tabularnewline
        \textit{Ours} (CLIP)    & 0.565 & 1.846 & 59.93  \tabularnewline
	\bottomrule
	\end{tabular}}
        \vspace{-0.02in}
	\caption{Descriptions as text queries} \label{subtab:description}
\end{subtable}

\begin{subtable}{1.\linewidth}
	\centering
	\setlength{\tabcolsep}{4.1mm}{
	\begin{tabular}{lccc} 
	\toprule 
	\multicolumn{1}{c}{\textsc{Method}} & TMD ($\times 10^1 $) $\uparrow$ & FPD $\downarrow$ & Acc. $\uparrow$ \tabularnewline 
	\midrule
        AutoSDF \cite{mittal2022autosdf} & 0.752 & 5.84 & 41.09 \tabularnewline 
	CLIP-Forge \cite{sanghi2022clip} & 0.961 & \textbf{4.14} & 55.00  \tabularnewline 
	\textit{Ours} (CLIP)  & \textbf{1.246} & 5.39 & \textbf{60.87}  \tabularnewline 
	\bottomrule
	\end{tabular}}
        \vspace{-0.02in}
	\caption{Prompts as text queries} \label{subtab:prompt}
\end{subtable}
\vspace{-0.2in}
\caption{Quantitative results of text-guide generation. }
\label{tab:text_generation}
\vspace{-0.2in}
\end{table}

\begin{figure}
\begin{centering}
\includegraphics[scale=0.47]{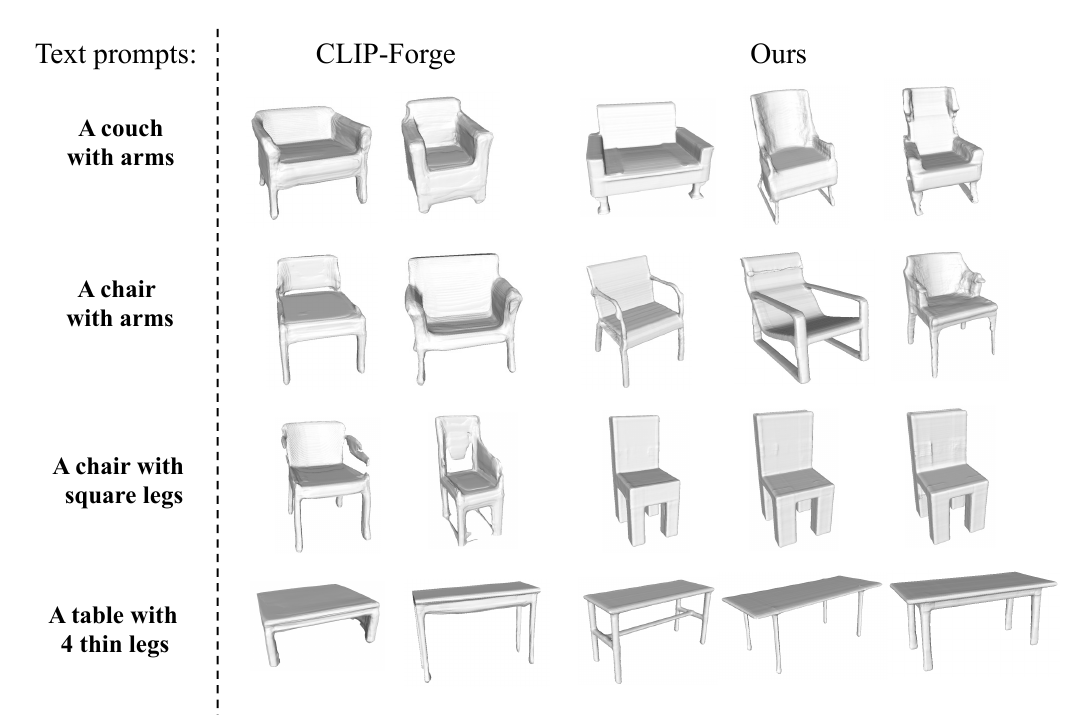}
\vspace{-0.1in}
\caption{Qualitative comparisons of text-guide generation, using prompts as text queries. Models are trained on Text2Shape dataset which only includes two categories, \textit{i.e.}, chair and table. \label{fig:text1}}
\vspace{-0.05in}
\end{centering}
\end{figure}

\noindent \textbf{Metrics and Settings.} All models are trained on the train split of Text2Shape dataset with their official codes. We evaluate our model with two types of text queries: (1) description: the test split of Text2Shape; (2) prompts: customized short phrases provided by \cite{sanghi2022clip} containing attributes for chair and table. We additionally use Accuracy (Acc.) \cite{sanghi2022clip} to measure the fidelity. The Accuracy is calculated by a pre-trained PointNet \cite{qi2017pointnet} classifier on ShapeNet.

\noindent \textbf{Results Analysis.} We report quantitative results of text-guide generation in Tab.~\ref{tab:text_generation}. Argus3D achieves promising results on TMD, MMD and Acc, showing good generalization performance across different conditional generation tasks.
Qualitative results are illustrated in Fig.~\ref{fig:text1}.
Furthermore, different from images, texts have a natural form of sequence. Each word is closely related to its context. Thereby, we discuss the ability of Argus3D to text-guide generation conditioned on sequence embeddings. Concretely, we adopt BERT and CLIP \footnote{In this case, we output features of all tokens instead of the last [END] token as the text embedding.} model to encode texts into a fixed length of sequence embeddings. From Tab.~\ref{subtab:description}, we find that using sequence embeddings indeed boosts the performance of Argus3D for text-guide generation tasks. Thanks to our proposed compact discrete representation with more tractable orders, our Argus3D can be easily adapted to different conditional forms for different tasks, thus improving the quality of the generated shapes.

\begin{table}                              
\centering  
\footnotesize{                              
\setlength{\tabcolsep}{0.5mm}   
\begin{tabular}{clcccccccc}            
\toprule                       
\multirow{2}{*}{\textsc{Category}} & \multicolumn{1}{c}{\multirow{2}{*}{\textsc{Method}}} &\multicolumn{2}{c}{MMD $\downarrow$}&\multicolumn{2}{c}{COV $\uparrow$}&\multicolumn{2}{c}{1-NNA $\downarrow$}&\multicolumn{2}{c}{ECD $\downarrow$}
\\  \cmidrule(lr){3-4} \cmidrule(lr){5-6} \cmidrule(lr){7-8} \cmidrule(lr){9-10}
   &  &CD  &EMD   &CD   &EMD   &CD   &EMD 	&CD   &EMD 			
\\  \midrule                                       
\multirow{5}{*}{Plane}      & GBIF \cite{ibing20213d} & 1439          & 1135          & 42.33          & 45.30           & 89.23          & 79.83          & 1563         & 1355         \\
 & AutoSDF \cite{autosdf}  & 4805          & 2481          & 26.57          & 31.32          & 80.66          & 79.30           & 4750         & 5101         \\
 & 3DILG \cite{zhang20223dilg}   & 1399          & 977          & 39.11          & 47.52          & 73.27          & 59.16           & 197         & 168         \\
 \cmidrule{2-10} 
      & Argus3D-B        & \textbf{1058} & 859 & 49.01          & 50.99          & 63.49          & 55.32          & 44           & 92           \\ 
      & Argus3D-H         & 1071           & \textbf{855}          & \textbf{50.25} & \textbf{53.71} & \textbf{57.80} & \textbf{49.01}  & \textbf{1}  & \textbf{68}  \\ \midrule
      
\multirow{5}{*}{Chair}      & GBIF \cite{ibing20213d} & 4252           & 2337          & 49.63          & 52.73          & 68.46          & 62.85          & 184          & 290          \\
 & AutoSDF \cite{autosdf}  & 5521          & 2716          & 34.71          & 37.67          & 81.39          & 76.29          & 3166         & 2991         \\
 & 3DILG \cite{zhang20223dilg}   & 5137          & 2596          &  41.72          & 42.31          & 65.98          & 64.05          & 323         & 351         \\
      \cmidrule{2-10}
      & Argus3D-B        & 4114          & 2262           & 49.63          & 51.85         & 59.90           & 56.94          & 80           & 183          \\ 
      & Argus3D-H    & \textbf{3062} & \textbf{1881} & \textbf{54.45}  & \textbf{54.21} & \textbf{50.83} & \textbf{50.18}  & \textbf{2}   & \textbf{87}  \\ \midrule
      
 \multirow{5}{*}{Table}     & GBIF \cite{ibing20213d} & 3481          & 2101           & 50.89          & 53.86          & 66.67          & 62.99          & 98           & 457          \\
 & AutoSDF \cite{autosdf}  & 4805          & 2481          & 26.57          & 31.32          & 80.66          & 79.30           & 4750         & 5101         \\
  & 3DILG \cite{zhang20223dilg}  & 5745          & 2669          & 28.98          & 32.78          & 75.89          & 74.47           & 1709         & 1635         \\
      \cmidrule{2-10} 
      & Argus3D-B        & 3397          & 1980          & 48.64          & 53.86          & 56.17          & 53.68          & 52           & 114          \\ 
      & Argus3D-H    & \textbf{3071} & \textbf{1867} & \textbf{55.63}  & \textbf{54.21} & \textbf{51.13} & \textbf{49.88}  & \textbf{1}   & \textbf{62}  \\ \midrule
      
 \multirow{5}{*}{Car}     & GBIF \cite{ibing20213d} & 1285          & 906          & 33.71          & 40.00             & 90.00             & 79.14          & 2453         & 1471         \\
  & AutoSDF \cite{autosdf}  & 1269          & 874          & 34.86          & 41.71          & 88.29          & 80.71          & 2312         & 1299         \\
    & 3DILG \cite{zhang20223dilg} & 1291          & 1040         & 26.86          & 31.43          & 81.43          & 75.71          & 847         & 741         \\
      \cmidrule{2-10} 
      & Argus3D-B        & 1193          & 870          & 36.45          & 37.38          & 73.77          & 72.43          & 1823         & 2232         \\ 
      & Argus3D-H     & \textbf{1100} & \textbf{834} & \textbf{43.43} & \textbf{41.71} & \textbf{68.57}    & \textbf{65.00} & \textbf{334} & \textbf{532} \\  \midrule
      
\multirow{5}{*}{AVG}        & GBIF \cite{ibing20213d} & 2614          & 1620          & 44.14          & 47.97          & 78.59          & 71.20          & 1075         & 894          \\
  & AutoSDF \cite{autosdf}  & 3398          & 1905          & 28.62          & 32.94          & 84.93          & 81.48          & 3669         & 3535         \\
   & 3DILG\cite{zhang20223dilg}   & 3393          & 1821          & 34.18          & 38.51          & 74.14          & 68.35          & 769         & 724         \\
        \cmidrule{2-10}
        & Argus3D-B        & 2441          & 1493          & 45.93          & 48.52          & 63.33          & 59.59          & 500          & 655          \\  
        & Argus3D-H      & \textbf{2076} & \textbf{1359} & \textbf{50.94} & \textbf{50.96} & \textbf{57.08} & \textbf{53.52} & \textbf{85} & \textbf{187} \\
\hline                    
\end{tabular}} 
\vspace{-0.05in}
\caption{Result comparisons of our large model on class-guide generation task. Best results are highlighted in \textbf{bold}.}
\label{tab:class}   
\vspace{-0.2in}
\end{table}

\begin{figure}[t]
    \centering
    \includegraphics[width=0.75\linewidth]{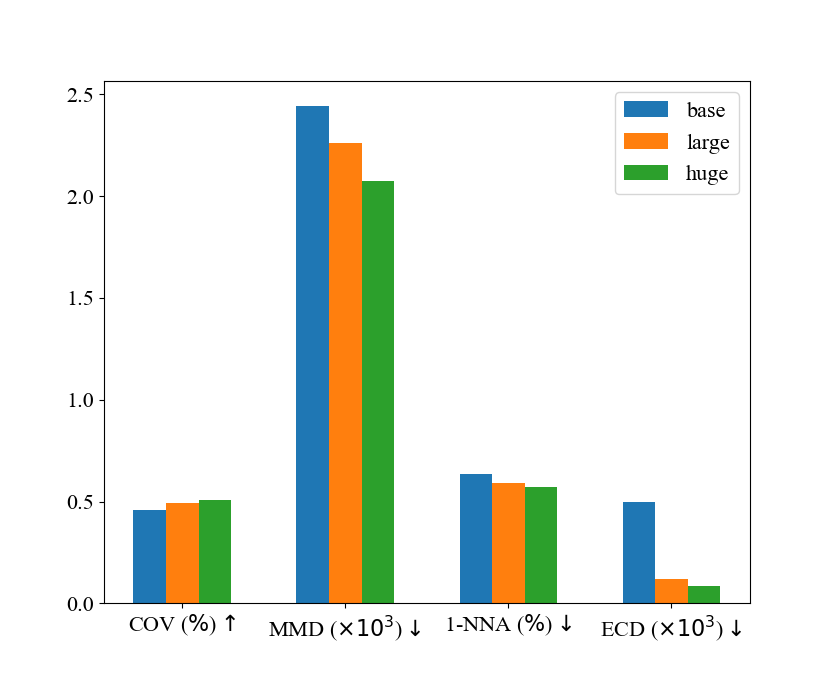}
    \vspace{-0.1in}
    \caption{Quantitative comparisons of our Argus3D with different scales. Models are trained for class-guide generation on ShapeNet.}
    \label{fig:cmp}
\vspace{-0.1in}
\end{figure}

\begin{figure}
    \centering
    \includegraphics[width=\linewidth]{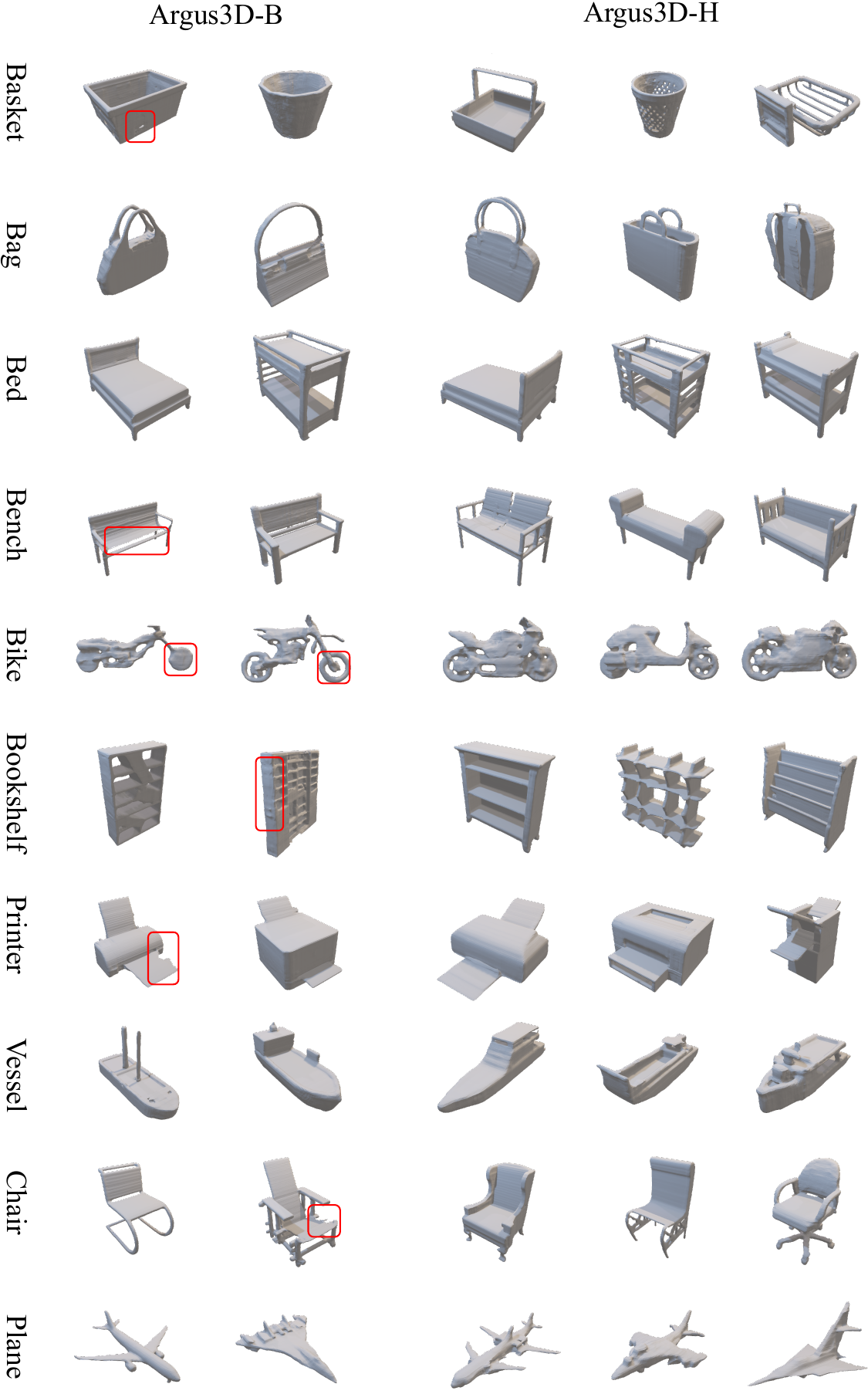}
    \caption{Quantitative results of class-guide generation for different Argus3D models. Red boxes indicate small flaws or missing details in the generated shapes.}
    \label{fig:vis-class-huge}
\end{figure}

\subsection{Generation with Larger Scale \label{sec:scale}}
As discussed in Sec.~\ref{sec:large_model}, taking advantage of our compact discrete representation and vanilla transformer structure, we scale up our Argus3D, from a base model of 100 million to a huge model of 3.6 billion, in search of stronger generation capacity endowed by larger parameters. 
In this section, we thus focus on investigating the efficacy of Argus3D at different scales, as well as the impact of our large-scale 3D dataset Objaverse-Mix. 
We evaluate models on two conditional 3D shape generation tasks, \textit{i.e.}, class-guide and image-guide, taking into account the challenge and reliability of tasks. 
The experimental settings and metrics are the same as the corresponding task above, unless are specifically explained.

\noindent \textbf{Benefit from Larger Models.}
We use the same ShapeNet dataset to train three Argus3D models at the base, large, and huge scales respectively, for the task of class-guide generation. The parameters of three models are listed in Tab.~\ref{tab:para}. In particular, they all share weights of the first-stage model.
Unlike Sec.~\ref{sec:class_generation}, we employ Chamfer Distance (CD) and Earth Mover's Distance (EMD) as distance metrics to evaluate performance for efficiency.
These metrics offer different perspectives on the model's ability to generate accurate and diverse shapes. 

We first discuss the effect of model parameters on the generation quality. Figure~\ref{fig:cmp} demonstrates the results of our three Argus3D models. As expected, with the increase of model parameters, the performance of four metrics is substantially improved. In addition, we compare our model with state-of-the-art approaches in Tab.~\ref{tab:uncond}. An observation similar to Tab.~\ref{tab:class_cond} can be found that our base model Argus3D-B already outperforms those competitors by a significant margin. And our largest model Argus3D-H can further boost the performance. It strongly suggests the superiority of our method for 3D shape generation at capacity and scalability.
The qualitative comparison is shown in Fig.~\ref{fig:vis-class-huge}, highlighting the capability of our model to generate shapes with more intricate details and complex
structures, thanks to the power of more learnable parameters.

\begin{figure}
    \centering
    \includegraphics[width=0.9\linewidth]{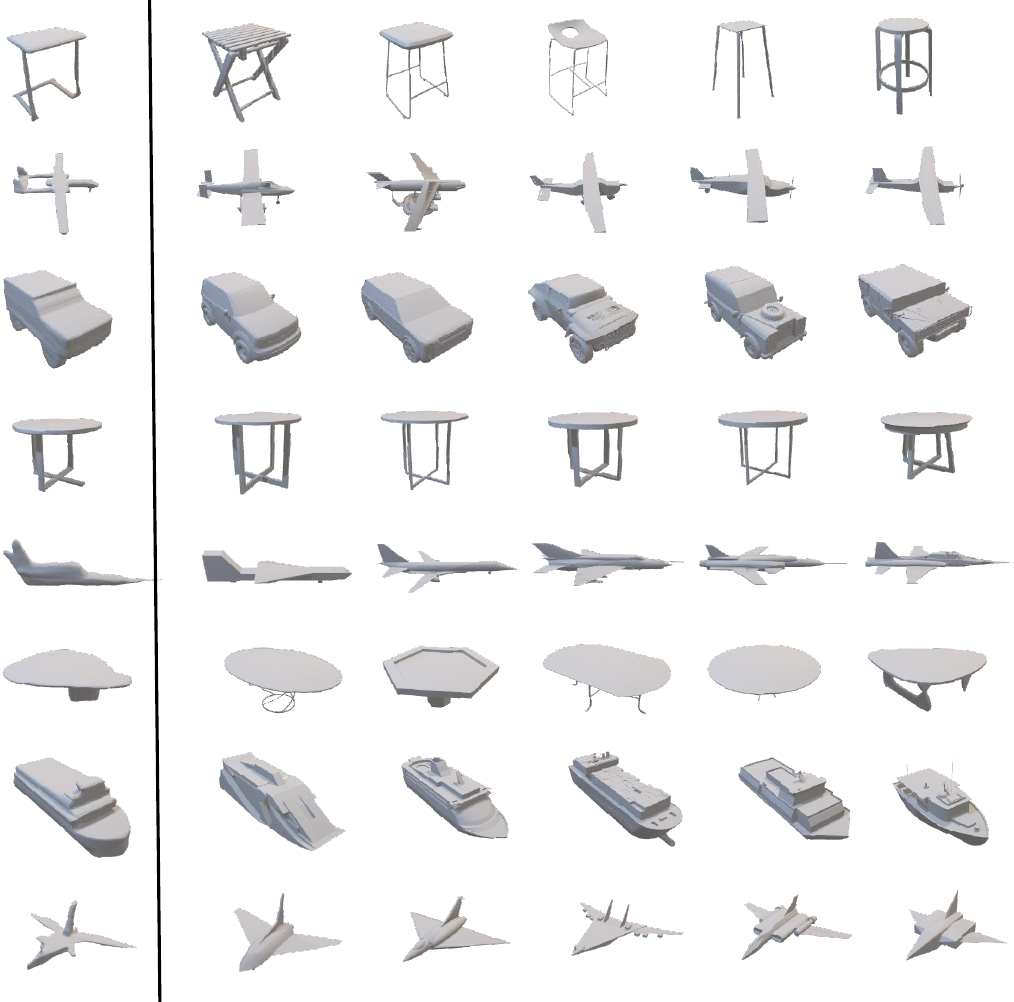}
    \vspace{-0.05in}
    \caption{Visualizations of the most similar shapes of generated ones in the training set.}
    \label{fig:cloest-shape}
\vspace{-0.05in}
\end{figure}

One possible concern is raised that a large number of parameters may cause the model to overfit, resulting in the model memorizing shapes rather than generalizing them. To verify it, we retrieve ground-truth shapes closest to our generated ones in the training set via Chamfer distance. Visualizations in Fig.~\ref{fig:cloest-shape} show five nearest neighbors with the similarity decreasing from left to right, and indicate that our Argus3D-H indeed learns the underlying distribution of 3D shapes rather than simply memorizing the training set to reconstruct them.

\begin{table}
\centering
\footnotesize
\setlength{\tabcolsep}{1.8mm}{
\begin{tabular}{lcccc} 
\toprule
\multicolumn{1}{c}{\textsc{Method}} & Pre. & TMD ($\times 10^2 $) $\uparrow$ & MMD ($\times 10^3$) $\downarrow$ & FPD $\downarrow$ \tabularnewline 
\midrule
AutoSDF \cite{autosdf} & $\scalebox{0.75}{\usym{2613}}$ & 2.657 & 2.137 & 15.262 \tabularnewline 
CLIP-Forge \cite{sanghi2022clip} & $\scalebox{0.75}{\usym{2613}}$ & 2.858 & 1.926 & 8.094   \tabularnewline 
\midrule
Argus3D-B (CLIP) & $\scalebox{0.75}{\usym{2613}}$ & 4.274 & 1.590 & 1.680  \tabularnewline 
Argus3D-H (CLIP) & $\checkmark$ & \textbf{5.136} & \textbf{1.338} & \textbf{0.774}  \tabularnewline 
\bottomrule
\end{tabular}}
\vspace{-0.05in}
\caption{Quantitative comparisons of our large model on image-guide generation. `Pre.' means whether using our Objaverse-Mix for pre-training.
} \label{tab:image_generation_large}
\end{table}

\noindent \textbf{Benefit from Larger datasets.}
To evaluate the efficacy of our proposed large-scale 3D dataset, we train Argus3D-H for image-guide generation. 
Concretely, the first-stage model is trained using occupancy labels from the Objaverse-Mix dataset, and then the second-stage training is performed with rendered multi-view images of 3D shapes. We further fine-tune the second-stage model on ShapeNet dataset, since the testing set is split from its 13 categories. Other experimental settings are the same as described in Sec.~\ref{sec:image_generation}.

Results in Tab.~\ref{tab:image_generation_large} showcase that Argus3D-H beats both our base model and other competitors. In particular, it has a significant advantage on TMD and FPD metrics. Thanks to the extensive prior knowledge from the large-scale Objaverse-Mix dataset, we can synthesize 3D shapes that are not only faithful to the condition images, but further enhance the diversity.
Additionally, we conducted tests on the Argus3D-Huge model using novel images generated by DALL·E, as depicted in Fig.~\ref{fig:dalle}. 
Although the category of synthesized images is seen before, Argus3D is capable of generating a wide range of shapes from unseen shapes, such as the shape of an avocado. Interestingly, by attaching a texture model, the synthesized shapes are more vivid and faithful to the conditional images.
It not only provides another perspective of text-guide 3D shape generation, but also
effectively demonstrates the superiority of our method in terms of robustness and versatility, affirming again the good properties of fidelity and diversity of our generated 3D shapes.

\subsection{Ablation Study\label{sec:more_results}}

Last but not least, we provide in-depth studies to discuss our motivations, and dissect the efficacy of our proposed designs and real-world applications.
Without loss of generality, all experiments are conducted with the Argus3D-Base model.

\begin{figure}[t]
    \centering
    \includegraphics[width=\linewidth]{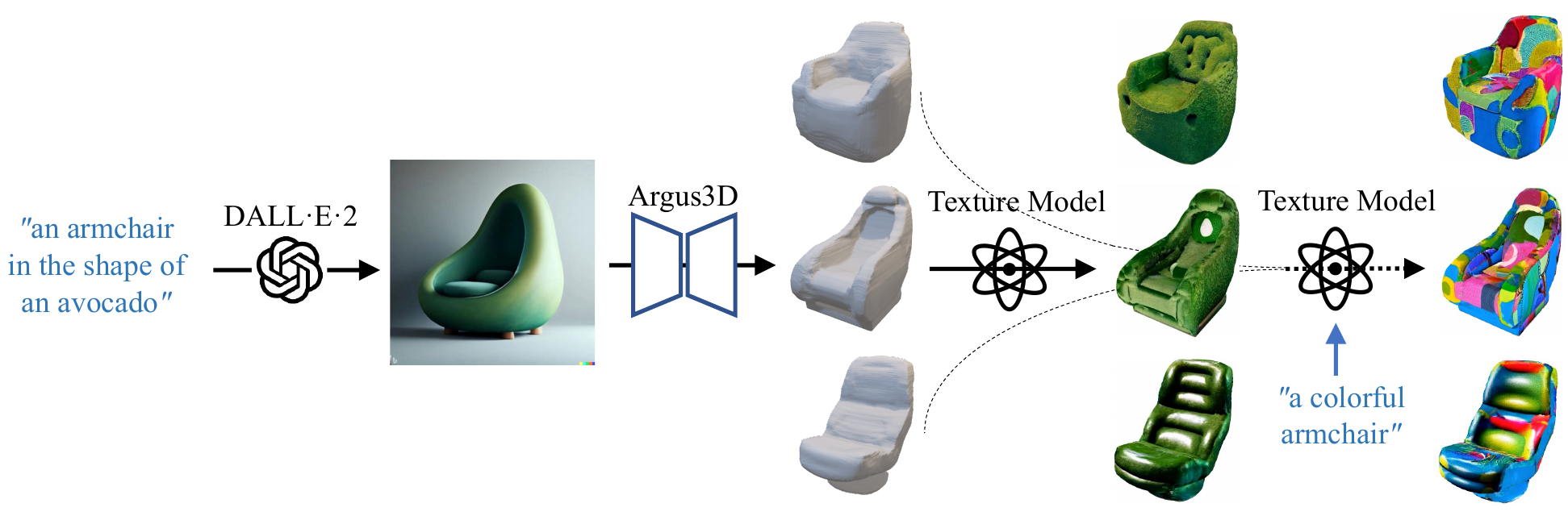}
    \caption{Image-guide generation based on DALL·E·2 generated images. 
    Argus3D is capable of generating a wide range of shapes from unseen images. These shapes can be further enhanced with textures created by a texture model, which utilizes text prompts from DALL·E 2. Additionally, the use of various text prompts enables the generation of new and unique textures.}
    \label{fig:dalle}
\end{figure}

\begin{table}
\centering
\footnotesize 
\setlength{\tabcolsep}{1.5mm}{
\begin{tabular}{cclcccc}
\toprule
\multirow{2}{*}{\textsc{Metrics}} &  \multicolumn{2}{c}{\multirow{2}{*}{\textsc{Methods}}} & \multicolumn{4}{c}{\textsc{Categories}}  \tabularnewline
\cmidrule{4-7}
 & \multicolumn{1}{c}{} & & Plane & Rifle & Chair & Car \tabularnewline 
\midrule
\multirow{5}{*}{ECD $\downarrow$} & \multirow{3}{*}{Tri-Plane}
& Iter-A & 744 & 405 & 4966 & 3599 \tabularnewline
& & Iter-B & 3501 & \textbf{36} & 1823 & 4735 \tabularnewline
& & Iter-C & 3098 & 282 & 4749 & 3193 \tabularnewline
\cmidrule{2-7}
& \multirow{2}{*}{Vector (\textit{ours})} & Row-Major & 236  & 65 & \textbf{27} & \textbf{842} \tabularnewline
& & Col-Major & \textbf{205} & 79 & 102 & 980 \tabularnewline

\midrule
\multirow{5}{*}{1-NNA $\downarrow$} & \multirow{3}{*}{Tri-Plane}
& Iter-A & 73.67 & 68.35 & 78.15 & 87.16\tabularnewline
& & Iter-B & 83.37 & \textbf{56.54} & 70.92 & 87.42 \tabularnewline
& & Iter-C & 81.83 & 65.61 & 78.38 & 88.53 \tabularnewline
\cmidrule{2-7}
& \multirow{2}{*}{Vector (\textit{ours})} & Row-Major & \textbf{59.95}  & 57.28 & \textbf{57.31} & \textbf{76.58} \tabularnewline
& & Col-Major & 62.48 & 57.70 & 58.38  & 78.09 \tabularnewline

\midrule
\multirow{5}{*}{COV $\uparrow$} & \multirow{3}{*}{Tri-Plane}
& Iter-A & \textbf{81.70} & 75.10 & 79.33 & 65.31 \tabularnewline
& & Iter-B & 74.16 & 75.52 & \textbf{82.95} & 63.97\tabularnewline
& & Iter-C & 71.32 & \textbf{76.69} & 78.89 & 72.25\tabularnewline
\cmidrule{2-7}
& \multirow{2}{*}{Vector (\textit{ours})} & Row-Major & 79.11 & 74.26 & 80.81 & \textbf{73.25} \tabularnewline
&  & Col-Major & 77.87 & 73.52 & 81.03 & 71.31\tabularnewline

\midrule
\multirow{5}{*}{CovT $\uparrow$} & \multirow{3}{*}{Tri-Plane}
& Iter-A & 43.51 & 41.56 & 23.10 & 50.30 \tabularnewline
& & Iter-B & 26.57 & 49.78 & 35.50 & 49.43 \tabularnewline
& & Iter-C & 30.28 & 41.35 & 24.94 & 51.23 \tabularnewline
\cmidrule{2-7}
& \multirow{2}{*}{Vector (\textit{ours})} & Row-Major & \textbf{45.12} & \textbf{55.27} & \textbf{49.82} & \textbf{56.64} \tabularnewline
& & Col-Major & 44.87 & 53.58 & 48.93 & 56.63 \tabularnewline

\midrule
\multirow{5}{*}{MMD $\downarrow$} & \multirow{3}{*}{Tri-Plane}
& Iter-A & 3237 & 3962 & 3392 & 1373 \tabularnewline
& & Iter-B & 3860 & 3624 & 3119 & 1404 \tabularnewline
& & Iter-C & 3631 & 3958 & 3430 & 1385 \tabularnewline
\cmidrule{2-7}
& \multirow{2}{*}{Vector (\textit{ours})} & Row-Major & 3124 & 3628  & \textbf{2703} & \textbf{1213} \tabularnewline
& & Col-Major & \textbf{3102} & \textbf{3623} & 2707 & 1214 \tabularnewline
\bottomrule
\end{tabular}}
\vspace{-0.05in}
\caption{The effect of flattening order on different discrete representations. Here, we take unconditional generation as an example and train one model per class. 
Please find statistic and qualitative analyses in Fig.~\ref{fig:order_hist} and \ref{fig:diff_view}, respectively.
\label{tab:uncond}}
\vspace{-0.1in}
\end{table}

\begin{figure*}
\includegraphics[scale=0.21]{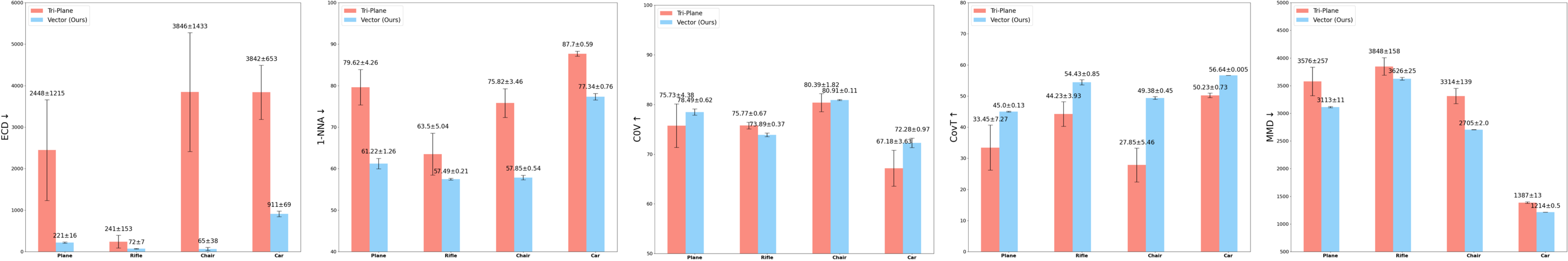}
\tabularnewline
\vspace{-0.05in}
\caption{Statistic analysis of the effect of flattening order. We report mean and standard deviation as histograms and error bars, respectively.
Best viewed in color and zoomed in.
\label{fig:order_hist}}
\vspace{-0.1in}
\end{figure*}

\begin{figure}
\begin{centering}
\includegraphics[scale=0.47]{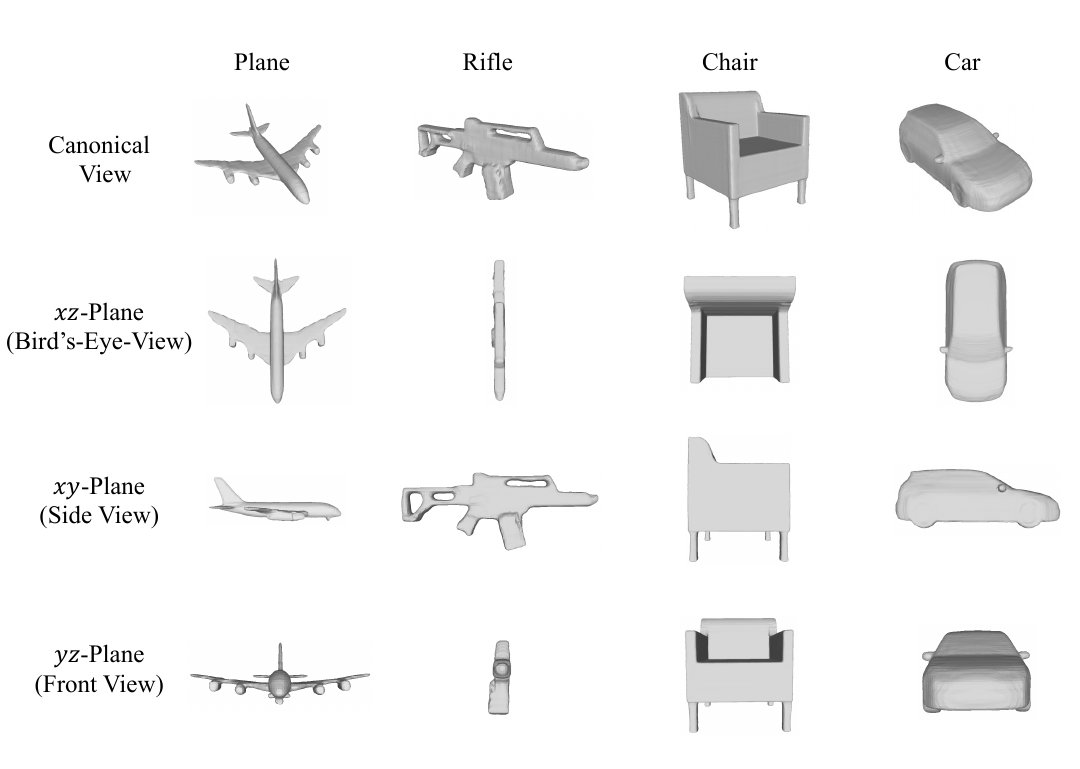}  
\tabularnewline
\vspace{-0.1in}
\caption{Visualizations of projected shapes on three planes.
\label{fig:diff_view}}
\vspace{-0.2in}
\end{centering}
\end{figure}

\noindent \textbf{Effect of Flattening Orders.} We first investigate how the flattening order affects the quality of shape generation. This study takes tri-planar representation as an example (`Tri-Plane' for short), since \textit{our improved discrete representation (`Vector') can naturally degenerate into `Tri-Plane' by removing the proposed coupling network}. As aforesaid in Fig.~\ref{fig:flatten_order}, different flattening orders will affect different auto-regressive generation orders of the three planes. Quantitatively, we consider three variants to learn joint distributions of tri-planar representation without loss of generality, Iter-A: $p\left(\mathbf{z}\right)=p\left(\mathbf{z}^{xz}\right)\cdot p\left(\mathbf{z}^{xy}|\mathbf{z}^{xz}\right)\cdot p\left(\mathbf{z}^{yz}|\mathbf{z}^{xz},\mathbf{z}^{xy}\right)$, Iter-B: $p\left(\mathbf{z}\right)=p\left(\mathbf{z}^{xy}\right)\cdot p\left(\mathbf{z}^{xz}|\mathbf{z}^{xy}\right)\cdot p\left(\mathbf{z}^{yz}|\mathbf{z}^{xy},\mathbf{z}^{xz}\right)$ and Iter-C:
$p\left(\mathbf{z}\right)=p\left(\mathbf{z}^{yz}\right)\cdot p\left(\mathbf{z}^{xz}|\mathbf{z}^{yz}\right)\cdot p\left(\mathbf{z}^{xy}|\mathbf{z}^{yz},\mathbf{z}^{xz}\right)$.

Results are presented in Tab.~\ref{tab:uncond} and Fig.~\ref{fig:order_hist}. As observed, different orders have a significant impact on performance, resulting in a large value of standard deviation. For instance, Iter-A achieves a better result on Plane category (Iter-A: 73.67 \textit{vs.} Iter-B: 83.37 on 1-NNA), while for Rifle, it prefers the order of Iter-B (Iter-B: 56.54  \textit{vs.} Iter-A: 68.35 on 1-NNA).
We attempt to explain this phenomenon by visualizing projected shapes on three planes. As illustrated in Fig.~\ref{fig:diff_view}, for Plane category (First Column), the projection onto the $xy$-plane provides limited shape information, while the $xz$-plane reveals more representative geometries of the aircraft.
We agree that if the $xz$-plane that contains more shape information is generated first, the generation of subsequent planes may be much easier.
Consequently, it is beneficial for Iter-A to generate more faithful 3D shapes than Iter-B. In contrast, Rifle and Chair exhibit more details on $xy$-plane, so the autoregressive order of Iter-B yields better results for these two categories. In addition, we notice that Car has a relatively simple shape, \textit{e.g.}, a cuboid, leading to similar impacts on the generation quality for different flattening orders.

\begin{figure}
\begin{centering}
\includegraphics[scale=0.38]{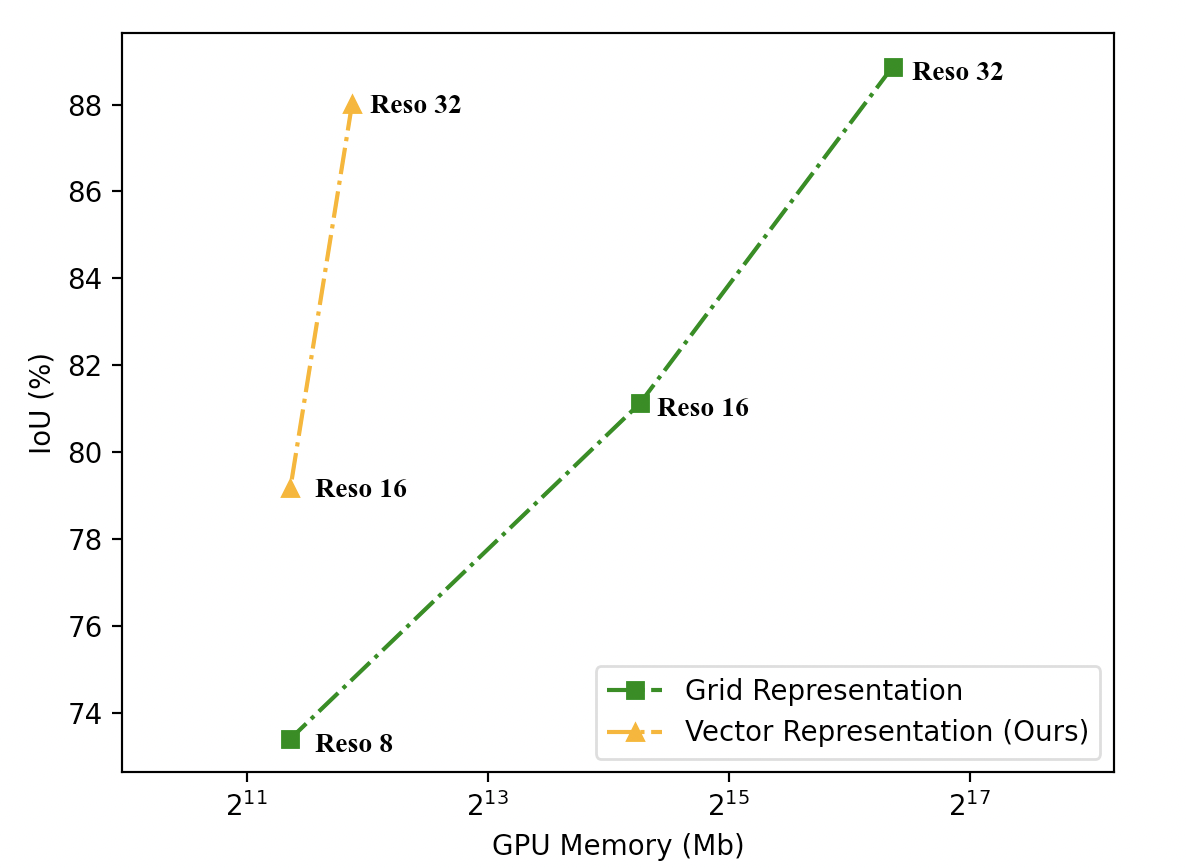}  
\tabularnewline
\vspace{-0.08in}
\caption{Comparisons of the first-stage IoU (\%) accuracy and the second-stage memory cost (Mb) with different resolutions and discrete representations. The memory cost is calculated with a batch size of 1.
\label{fig:grid_plane}}
\vspace{-0.15in}
\end{centering}
\end{figure}

Next, We evaluate the advantage of our improved discrete representation in terms of efficacy and efficiency.
We explore two variants of our full model by flattening coupled feature maps into vectors with row-major or column-major order. 
In Tab.~\ref{tab:uncond}, our proposed method achieves similar well performance even with different serialization orders. 
Figure~\ref{fig:order_hist} shows that our standard deviation is significantly lower than `Tri-Plane', demonstrating 
the robustness of our improved representation to generation orders. 
The proposed coupling network has facilitated AR learning by introducing more tractable order.
Additionally, the overall quality of synthesized shapes for all categories is balanced and excellent across all metrics, indicating the superiority of our design.

Furthermore, we also investigate the advantage of low computation overhead. We use `Vector' to denote our design since we apply vector quantization to latent vectors, and `Grid' refers to the baseline method that applies vector quantization to volumetric grids. Figure~\ref{fig:grid_plane} compares the performance of IoU at the first stage and the corresponding memory cost in the second stage. Since we cannot afford the training of transformers with volumetric grid representations, as an alternative, we report the first-stage IoU accuracy for comparisons. From Fig.~\ref{fig:grid_plane}, two conclusions can be drawn. \textbf{(1)} The resolution $r$ of feature grids (`Grid' and `Vector') significantly affects the quality of reconstructed shapes. If $r$ is too small, it lacks the capacity to represent intricate and detailed geometries (Grid Reso-32: $88.87$ \textit{v.s} Grid Reso-16: $81.12$). However, the large $r$  will inevitably increase the computational complexity in the second stage, as the number of required codes explodes as $r$ grows (Grid Reso-32: $\geq 80$G  \textit{v.s} Grid Reso-16: $19.6$G).
\textbf{(b)} Our proposed `Vector' representation not only achieves comparable reconstruction results (Vector Reso-32: $88.01$ \textit{v.s} Grid Reso-32: $88.87$), but also significantly reduces the computation overhead (Vector Reso-32: $3.8$G \textit{v.s} Grid Reso-32: $\geq 80$G).

\begin{table}
\centering
\footnotesize 
\setlength{\tabcolsep}{1.6mm}{
\begin{tabular}{c|c|c|c|c|c} 
\toprule
\multirow{2}{*}{\textsc{Num.}} &\multirow{2}{*}{\textsc{Type}} & \multirow{2}{*}{\textsc{\#Entry}} & \multirow{2}{*}{\textsc{Reso.}} & \textsc{Stage 1} & \textsc{Stage 2} \tabularnewline 
\cline{5-6}
& & & & IoU $\uparrow$ & 1-NNA / ECD $\downarrow$  \tabularnewline 
\hline
\midrule
0 & Grid & \multirow{3}{*}{4096} & \multirow{3}{*}{32} & 88.87  & $\times$  \tabularnewline 
\cline{2-2} \cline{5-6}
1 & Tri-Plane &  & & 87.81 & 73.67 / 743 \tabularnewline 
\cline{2-2} \cline{5-6}
2 & \multirow{4}{*}{Vector} &  & & 88.01 & \bf{59.95 / 236} \tabularnewline 
\cline{3-6}
3 &  & 4096 & 16 & 79.17 & \multirow{3}{*}{-} \tabularnewline 
\cline{3-5}
4 &  & 2048 & \multirow{2}{*}{32} & 86.99 &  \tabularnewline 
\cline{3-3}\cline{5-5}
5 &  & 1024 &  & 86.57 &  \tabularnewline 
\bottomrule
\end{tabular}}
\vspace{-0.08in}
\caption{Ablation study of auto-encoder design choices. We report 1-NNA/ECD for plane category. `\textsc{Reso.}' means the resolution of feature map for vector quantization. `$\times$': cannot report due to extreme memory cost. `-': not report. 
\textit{Notably, without the coupling network, our method naturally degenerates into `Tri-Plane' representation.}
}
\label{tab:ablation_study}
\vspace{-0.05in}
\end{table}

\begin{figure}
\begin{centering}
\includegraphics[scale=0.36]{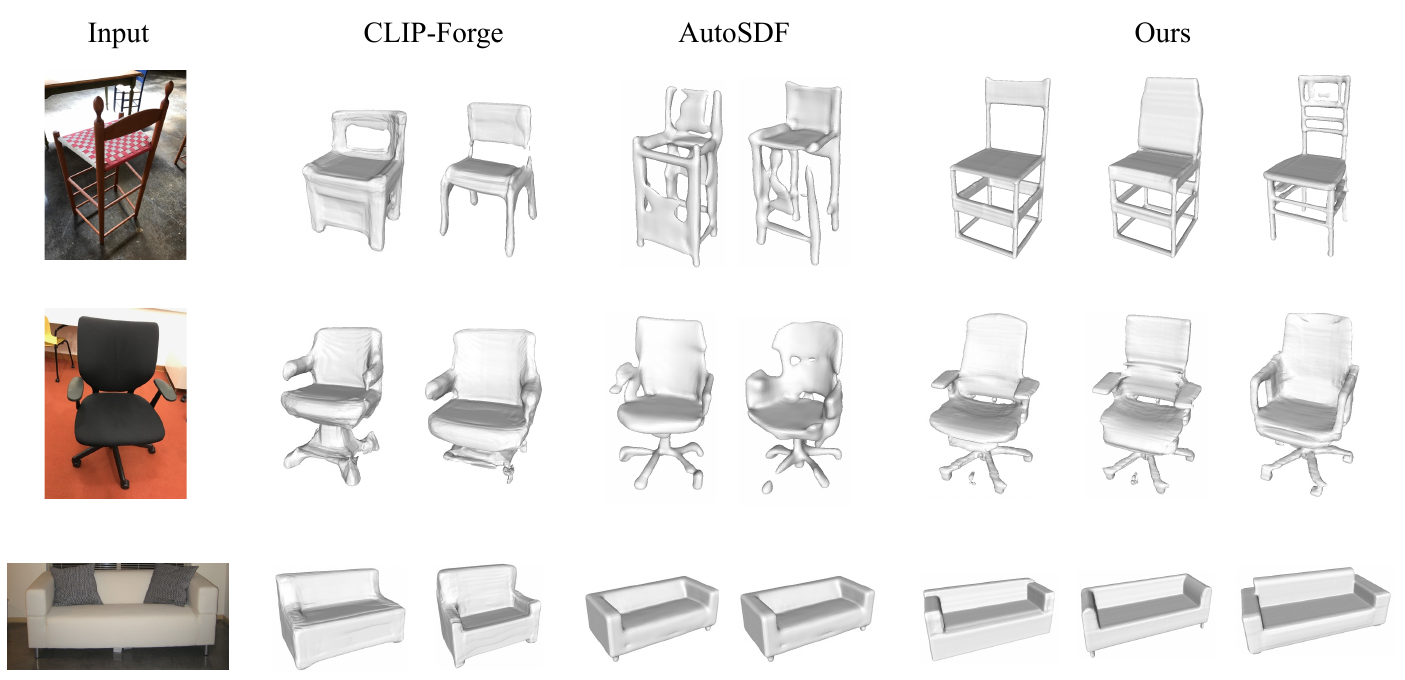}
\caption{Results of real-world image-guide generation. Samples are from Pix3D dataset.}
\label{fig:pix3d}
\end{centering}
\end{figure}

\noindent \textbf{Design Choices in Discrete Representation Learning.}
As a key contribution of this paper, we first discuss the efficacy of our improved discrete representation learning. 
Similarly, 'Tri-Plane' refers to baselines applying vector quantization to the three planes.
Results in Tab.~\ref{tab:ablation_study} show that `Grid' gets better IoU performance for shape reconstruction in the first stage, but fails to train transformers in the second stage due to the extreme long length of the sequence (\textit{e.g.}, $32^{3}$). 
In contrast, our model not only achieves comparable reconstruction results (\#0 \textit{vs.} \#2), but also outperforms 'Tri-Plane' by a large margin on generation quality (\#1 \textit{vs.} \#2). The latter shows inferior results due to `ambiguity of order' (see Suppl.)
It significantly proves the efficacy of our proposed coupling network, improving the plasticity and flexibility of the model.
We also explore different design choices in Tab.~\ref{tab:ablation_study}. By gradually increasing feature resolutions or the number of entries in the codebook, we achieve better performance on IoU. This observation is consistent with \cite{yu2021vector}, as the capacity of the discrete representation is affected by these two factors. 
A larger number of entries promises to further improve accuracy. Thus, we empirically double it for our large and huge models.

\begin{figure}
\begin{centering}
\includegraphics[scale=0.6]{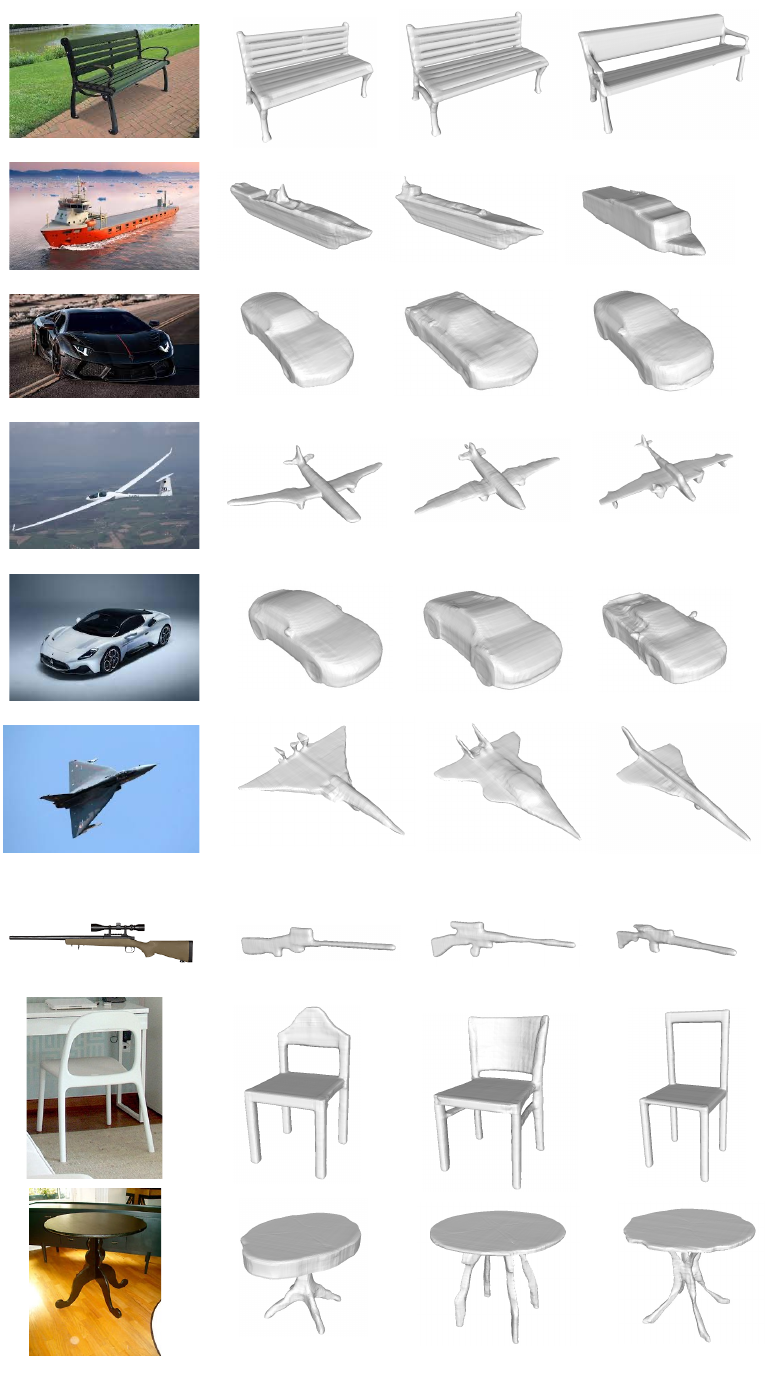}
\vspace{-0.15in}
\caption{Results of real-world image-guide generation. Images are randomly selected from the internet.}
\label{fig:real_world}
\vspace{-0.1in}
\end{centering}
\end{figure}

\begin{figure}
\begin{centering}
\includegraphics[scale=0.55]{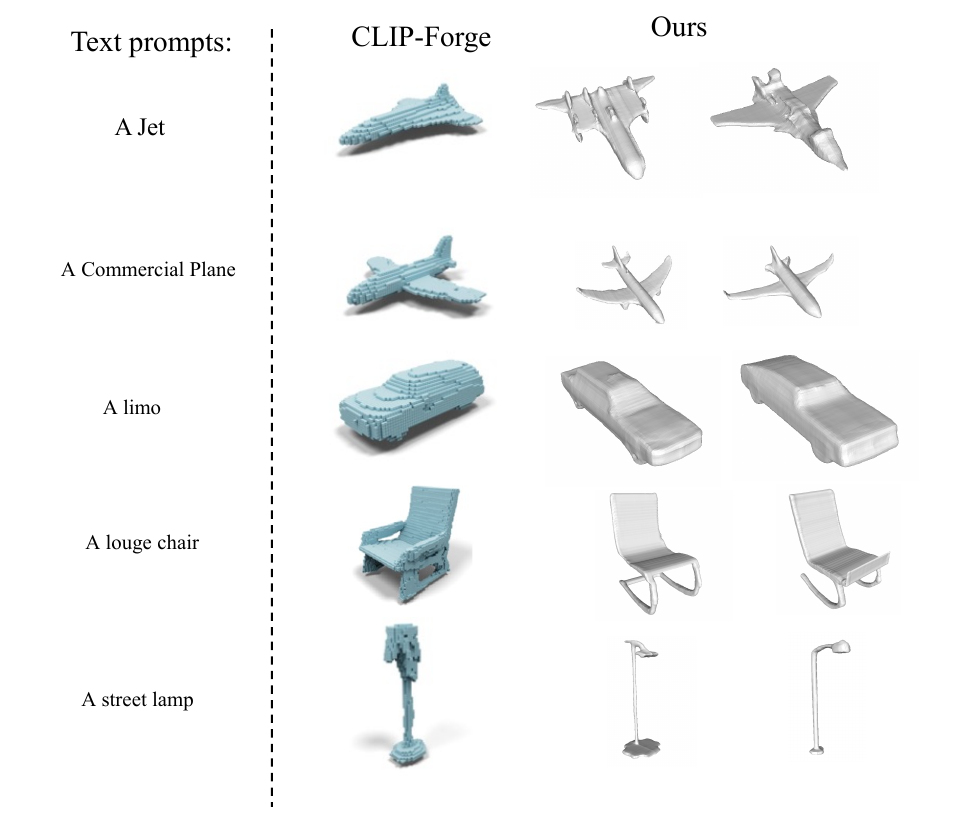}
\vspace{-0.25in}
\caption{Qualitative comparisons of zero-shot text-to-shape generation. Results of CLIP-Forge are reported in their paper.}
\label{fig:zero_shot_compare}
\vspace{-0.05in}
\end{centering}
\end{figure}

\begin{figure}
 \begin{centering}
\includegraphics[scale=0.43]{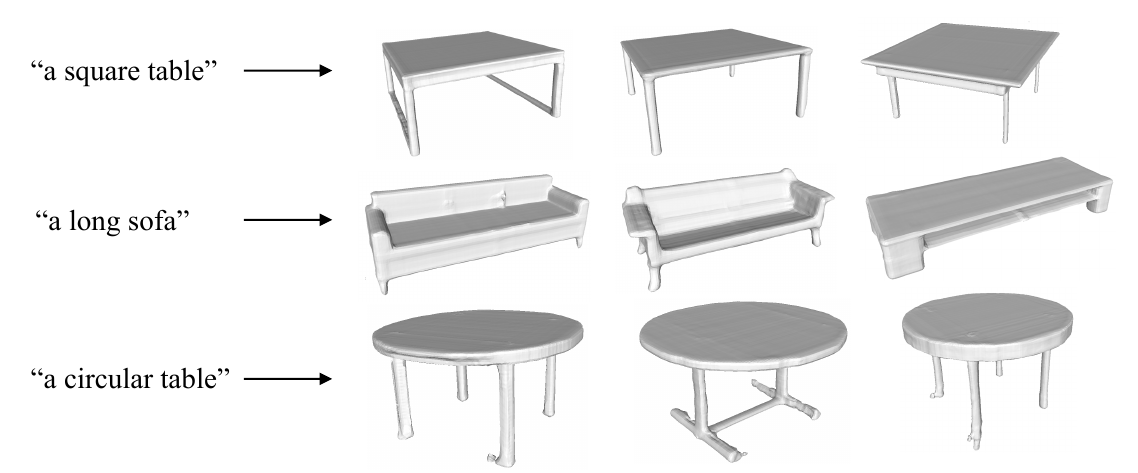}
\caption{More visualizations of zero-shot text-to-shape generation.}
\label{fig:zero_shot_vis}
\end{centering}
\end{figure}

\noindent \textbf{Image-guide Generation in Real-world.} We further investigate the generalizability of our model on real-world images. We use the model trained on ShapeNet as described in Sec.~\ref{sec:image_generation}, and download images from the internet as conditions. Fig.~\ref{fig:real_world} shows the qualitative results for different categories, \textit{i.e.}, plane, chair, car and table. Our model sensitively captures major attributes of objects in the image and produces shapes faithful to them. (see the first column to the left). 
Meanwhile, our synthesized samples enjoy the advantage of diversity by partially sticking to the images, such as the  types of wings and tail designs in the case of airplane images.

Figure~\ref{fig:pix3d} also shows the results of our model on Pix3D dataset (trained on ShapeNet, without any finetuning).
Compared with other competitors, Argus3D is capable of generating high-quality and realistic shapes that highly match the shape of objects in real-world images. It significantly highlights the strong generalization ability of our method.

\noindent \textbf{Zero-shot Text-to-shape Generation.}
Inspired by CLIP-Forge \cite{sanghi2022clip}, we utilize the CLIP model to achieve zero-shot text-to-shape generation. At training, we only use the rendered images of 3D shapes. At inference, we substitute image features with text features encoded by the CLIP model. Figure~\ref{fig:zero_shot_compare} and \ref{fig:zero_shot_vis} show our ability of zero-shot generation, 
where shape attributes are controlled with different prompts. 
The high-quality of synthesized shapes clearly
demonstrate the powerful versatility of our Argus3D
in 3D shape generation, showing great potential to real-world applications.

\section{Conclusion \label{sec:conclusion}}

We introduce an improved AR model for 3D shape generation. By projecting volumetric grids of encoded input shapes onto three axis-aligned orthogonal feature planes, which are then coupled into a latent vector, we reduce computational costs and create a more tractable order for AR learning. Our compact and tractable representations enable easy switching between unconditional and conditional generation with multi-modal conditioning inputs. Extensive experiments demonstrate that our model outperforms previous methods on multiple generation tasks.
By scaling up both the model parameters and dataset size, we have developed Argus3D-Huge, the largest 3D shape generation model currently available. Alongside it, we introduce the Objaverse-Mix dataset, a comprehensive collection containing multi-modal data. These advancements further enhance our model's performance, showcasing its robust capabilities in 3D shape generation.

\section*{Acknowledgement}
Yanwei Fu,  Xiangyang Xue, and Yinda Zhang are the corresponding authors.
This work is supported by China Postdoctoral Science Foundation (2022M710746), National Key R\&D Program of China (2021ZD0111102) and NSFC-62306046. 
Yanwei Fu is with the School of Data Science and MOE Frontiers Center for Brain Science, Fudan University. Yanwei Fu is also with Fudan ISTBI—ZJNU Algorithm Centre for Brain-inspired Intelligence, Zhejiang Normal University, Jinhua, China.

\bibliographystyle{IEEEtran}
\bibliography{egbib}

\ifCLASSOPTIONcompsoc
\else
\fi


\ifCLASSOPTIONcaptionsoff
  \newpage
\fi

\appendix
\section*{A. Model Architectures}

\begin{figure*}[t]
\begin{centering}
\includegraphics[scale=0.4]{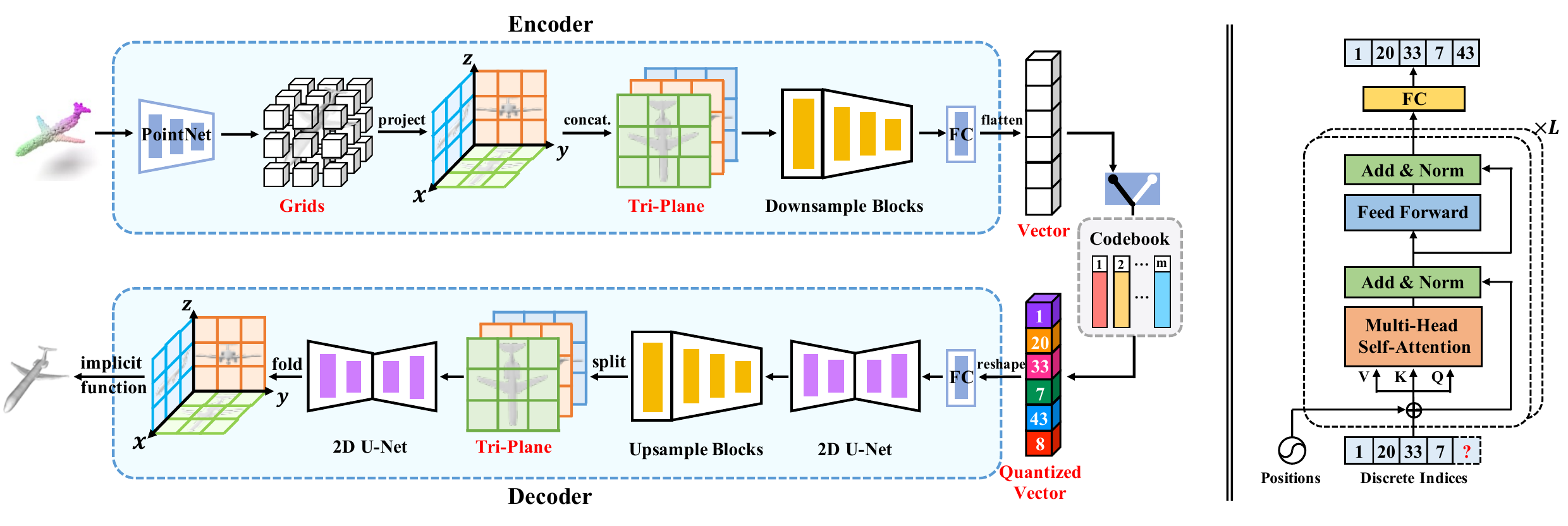}
\vspace{-0.05in}
\caption{The architectures of Argus3D, consisting of an auto-encoder at the first stage (left) and a transformer at the second stage (right). 
\label{fig:pipeline}}
\end{centering}
\vspace{-0.2in}
\end{figure*}

Our proposed framework Argus3D consists of a two-step procedure for 3D shape generation. The first step is an auto-encoder structure, aiming to learn discrete representations for input 3D shapes. And the second step introduces a transformer structure to learn the joint distribution of discrete representations. Below we will elaborate on the details of these two structures.

\noindent \textbf{Auto-encoder.} As shown in the left of Fig.~\ref{fig:pipeline}, the auto-encoder takes as input point clouds $\mathcal{P} \in \mathbb{R}^{n \times 3}$ with $n$ means the number of points, and outputs the predicted 3D mesh $\mathcal{M}$. More concretely, the encoder starts by feeding point clouds into a PointNet \cite{riegler2017octnetfusion} with local pooling, resulting in point features with dimensions $\mathbb{R}^{n \times 32}$. Then, we project points on three axis-aligned orthogonal planes with the resolution of $256$. Features of points falling into the same spatial grid cell are aggregated via mean-operation, so that input point clouds are represented as tri-planar features instead of volumetric features. To further improve the representation, we concatenate three feature planes and couple them with three convolution layers. Next, four stacked convolution layers are adopted to not only down-sample the feature resolution three times, but also highly encode and abstract the position mapping of each spatial grid in 3D space. Thus, the output has a tractable order to be serialized as a feature vector. Before we perform the vector quantization on the flattened outputs, we follow \cite{yu2021vector} to utilize the strategy of low dimensional codebook lookup, by squeezing the feature dimension from $256$ to $4$. Consequently, an arbitrary 3D shape can be represented with a compact quantized vector, whose elements are indices of those closest entries in the codebook.

The decoder is composed of two 2D U-Net modules and one symmetric upsample block. After reshaping the quantized vector and unsqueezing its feature dimension from $4$ to $256$, we apply a 2D U-Net module to complement each spatial grid feature with global knowledge. Subsequently, the same number of 2D convolution layers as the downsample block is appended to upsample the feature resolution back to $256$. Symmetric convolution layers further decouple it into tri-planer features. To further improve the smoothness between the spatial grids in each plane, we use the other shared 2D U-Net module to separately process tri-plane features. The structures of both 2D U-Net are in alignment with \cite{mescheder2019occupancy}. Finally, we build a stack of fully-connected residual blocks with $5$ layer, as an implicit function, to predict the occupancy probability of each query position.

\noindent \textbf{Transformer.} Benefiting from the compact discrete representation with a tractable order for each input 3D shape, we adopt a vanilla decoder-only transformer without any specific-designed module to learn the joint distributions among discrete codes. Our transformer consists of $T$ decoder layers, each of which has one multi-head self-attention layer and one feed-forward network. The decoder layer has the same structure as \cite{esser2021taming}, and is illustrated in the right of Fig.~\ref{fig:pipeline}. Specifically, we use a learnable start-of-sequence token ([SOS] token) to predict the first index of the discrete vector, and auto-regressively predict the next with the previous ones. For example, given an input containing the first $t$ indices along with one [SOS] token, we first use an embedding layer to encode them as features. Then, we feed them into several transformer layers. The position embedding is also added to provide positional information. At the end of the last transformer layer, we use two fully-connected layers to predict the \textit{logit} of each token. However, we only keep the last one which is a meaningful categorical distribution for the next ($t$+1)-th index.

\begin{table}[t]
\centering
\footnotesize{
\setlength{\tabcolsep}{4.5mm}{
\begin{tabular}{lll}
\toprule
\multicolumn{1}{c}{\textsc{Layer Name}} & \multicolumn{1}{c}{\textsc{Notes}} & \multicolumn{1}{c}{\textsc{Input Size}} \tabularnewline 
\midrule
\midrule
\textbf{Auto-encoder} & & \tabularnewline 
PointNet  &  & $n\times 3$ \tabularnewline 
Coupler & &  \tabularnewline
\quad ConvLayer & k3s1p1 & $256\times256\times3\times32$ \tabularnewline 
\quad ConvLayer & k3s1p1 & $256\times256\times96$ \tabularnewline 
\quad ConvLayer & k1s1p0 & $256\times256\times32$ \tabularnewline 
Downsampler & &  \tabularnewline 
\quad ConvLayer & k2s2p0 & $256\times256\times32$ \tabularnewline 
\quad ConvLayer & k2s2p0 & $128\times128\times64$ \tabularnewline 
\quad ConvLayer & k2s2p0 & $64\times64\times128$ \tabularnewline 
\quad ConvLayer & k1s1p0 & $32\times32\times256$ \tabularnewline 
Squeezer & k1s1p0 & $32\times32\times256$ \tabularnewline 
Quantizer & & $32\times32\times4$ \tabularnewline 
Unsqueezer & k1s1p0 & $32\times32\times4$ \tabularnewline 
2D U-Net & & $32\times32\times256$  \tabularnewline 
Upsampler & &  \tabularnewline 
\quad DeconvLayer & k3s1p1 & $32\times32\times256$ \tabularnewline 
\quad DeconvLayer & k3s1p1 & $64\times64\times128$ \tabularnewline 
\quad DeconvLayer & k3s1p1 & $128\times128\times64$ \tabularnewline 
\quad ConvLayer & k1s1p0 & $256\times256\times32$ \tabularnewline 
Decoupler & &  \tabularnewline
\quad ConvLayer & k3s1p1 & $256\times256\times32$ \tabularnewline 
\quad ConvLayer & k3s1p1 & $256\times256\times96$ \tabularnewline 
\quad ConvLayer & k1s1p0 & $256\times256\times96$ \tabularnewline 
2D U-Net & & $256\times256\times3\times32$  \tabularnewline
\midrule
\midrule
\textbf{Transformer} & & \tabularnewline 
Embedding Layer &  & $\left(1+\text{L}\right) \times 1$ \tabularnewline 
Decoder Layers $\times$ $T$ & &  \tabularnewline 
\quad Self-Attention & h & $\left(\text{K}+1+\text{L}\right) \times d$ \tabularnewline 
\quad Feed-Forward & m$4d$ & $\left(\text{K}+1+\text{L}\right) \times d$ \tabularnewline 
Head Layer & &  \tabularnewline 
\quad LinearLayer &  & $\text{L} \times d$ \tabularnewline 
\quad LinearLayer &  & $\text{L} \times d$ \tabularnewline 
\bottomrule
\end{tabular}}}
\vspace{-0.05in}
\caption{The detailed architecture of our framework. `k', `s' and `p' denote kernel size, stride and padding, respectively, in the convolution layer. `h' means the number of heads in multi-head self-attention layer, and `$T$' is the number of layers. The feature dimension $d$ in the transformer varies for different scales. $m$ stands for the dimension of the middle layer in the feed-forward network. `K' and `L' are the sequence length of conditioning inputs and discrete representation, and `1' indicates the length of [SOS] token. 
\label{tab:implentation}}
\vspace{-0.1in}
\end{table}

\noindent \textbf{Implementation Details.} Table~\ref{tab:implentation} summarizes all parameter settings for both auto-encoder and transformer structures. We apply them as default to all experiments unless otherwise stated. 
The parameter settings for Argus3D of different scales are shown in Tab.~\ref{tab:para}.
The number of entries $m$ in the codebook is 4096, 8192 and 8192 for Argus3D-Base, -Lager and -Huge, respectively. 
Lower triangular mask matrix is used in all multi-head self-attention layers to prevent information leakage, that is, the prediction of the current index is only related to the previous known indices. 
For various conditioning inputs, we adopt the most common way to encode them. 
For example, we use a learnable embedding layer to get the feature $\in \mathbb{R}^{1\times d}$ of each category.
Given partial point clouds, our proposed auto-encoder encodes them into discrete representations, which are fed into another embedding layer to get features with $d$ dimensions. We adopt pre-trained CLIP models to extract features $\in \mathbb{R}^{1\times512}$ for images or texts, and further use one fully-connected layer to increase the dimension from $512$ to $d$. All of the encoded conditioning inputs are simply prepended to [SOS] token via concatenation to guide the generation.

\section*{B. Training and Testing Procedures \label{sec:train_test}}

\noindent \textbf{Training:} 
All models are trained on at most eight A100 GPUs. 
For the first stage, we take dense point clouds with $n=30,000$ as input, and train the auto-encoder on ShapeNet (or Objaverse-Mix) dataset for a total 600k (or 1300k) iterations. The learning rate is set as 1e-4, and the batch size is $16$. Once trained, it is shared for all generation tasks.
For the second stage, we adopt the learning rate of 1e-4 for the base model, and 1e-5 for the large and huge models. The learning rate decay is manually adjusted for the training of Argus3D-H, and set it to 3e-6 and 1e-6 respectively when the loss approaches convergence. Due to the increase in the number of parameters, the batch size for three models of different scales is set to 8, 3 and 1, respectively.
It takes around 600k iterations for the base model on a single GPU card, and 3500k iterations for the huge model on eight GPU cards.

\noindent \textbf{Testing:} During inference, we first use the well-trained transformer to predict discrete index sequences with or without conditioning inputs. For each index, we sample it with the multinomial distribution according to the predicted probability, where only the top-\textit{k} indices with the highest confidence are kept for sampling. We progressively sample the next index, until all elements in the sequence are completed. Then, we feed the predicted index sequence into the decoder to get tri-planar features. Subsequently, we interpolate the feature of each point on a grid of resolution $128^3$ from tri-planar features, and adopt the implicit function to query the corresponding occupancy probability. Finally, the iso-surface of 3D shapes is extracted with the threshold of $0.2$ via Marching Cubes \cite{lorensen1987marching}.

\noindent \textbf{Inference Speed:} 
For our base model Argus3D-B, the wall time it takes to sample a single shape is roughly 14 seconds, and we can also generate 32 shapes in parallel in 3 minutes. For our Argus3D-H, the inference speed required to sample a single shape is about 50 seconds, due to the increased computation of more parameters and layers.

\section*{C. Visualization \label{sec:visual}}
We attach high-resolution visualizations of Figs.~\textcolor{red}{7}, \textcolor{red}{8}, \textcolor{red}{9}, and \textcolor{red}{15} for better views.

\begin{figure*}
\begin{centering}
\includegraphics[width=1.0\linewidth]{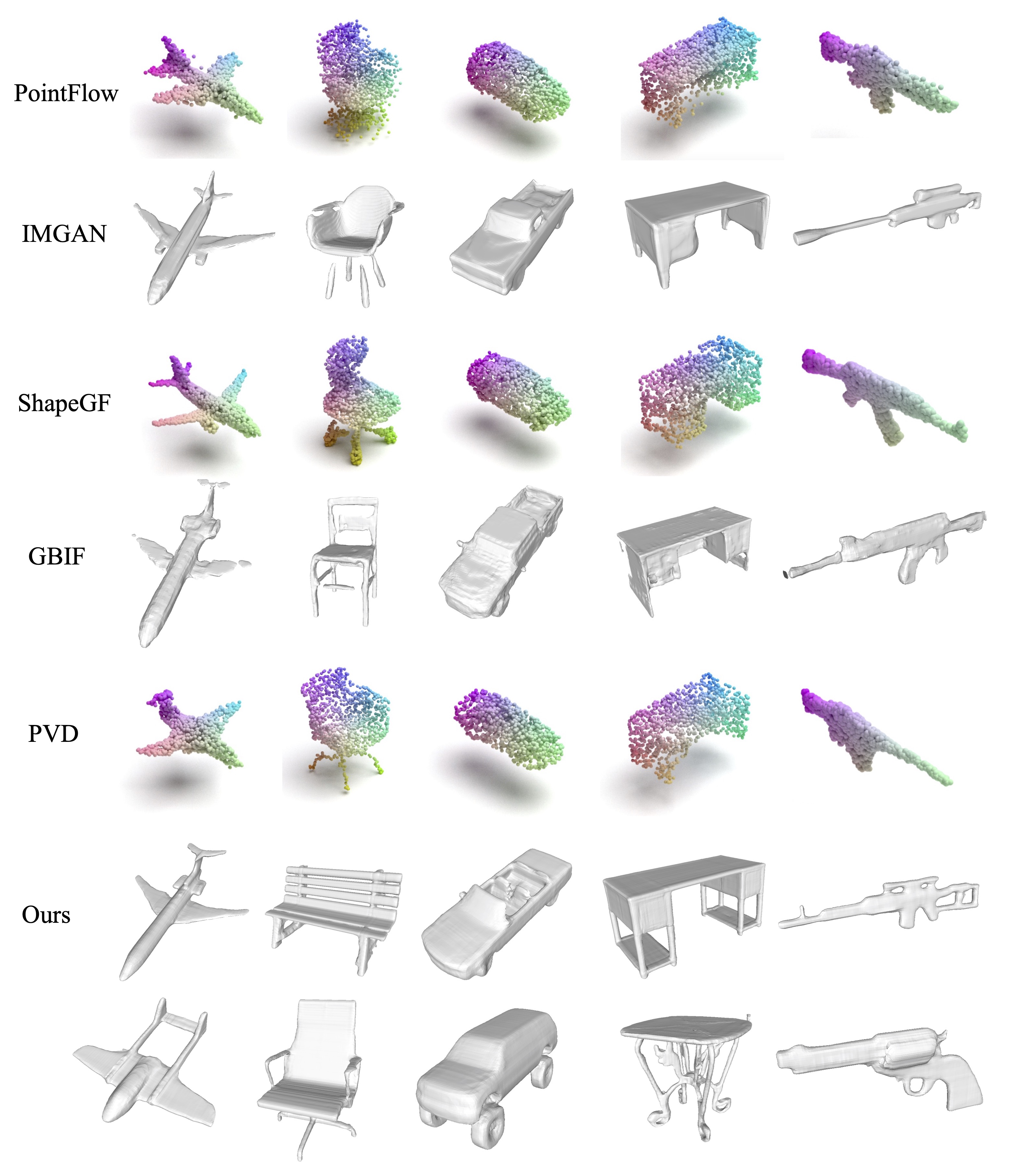}
\vspace{-0.1in}
\caption{Qualitative results of unconditional generation. \label{fig:unconditional} 
}
\vspace{-0.1in}
\end{centering}
\end{figure*}

\begin{figure*}[t]
\begin{centering}
\includegraphics[width=1.0\linewidth]{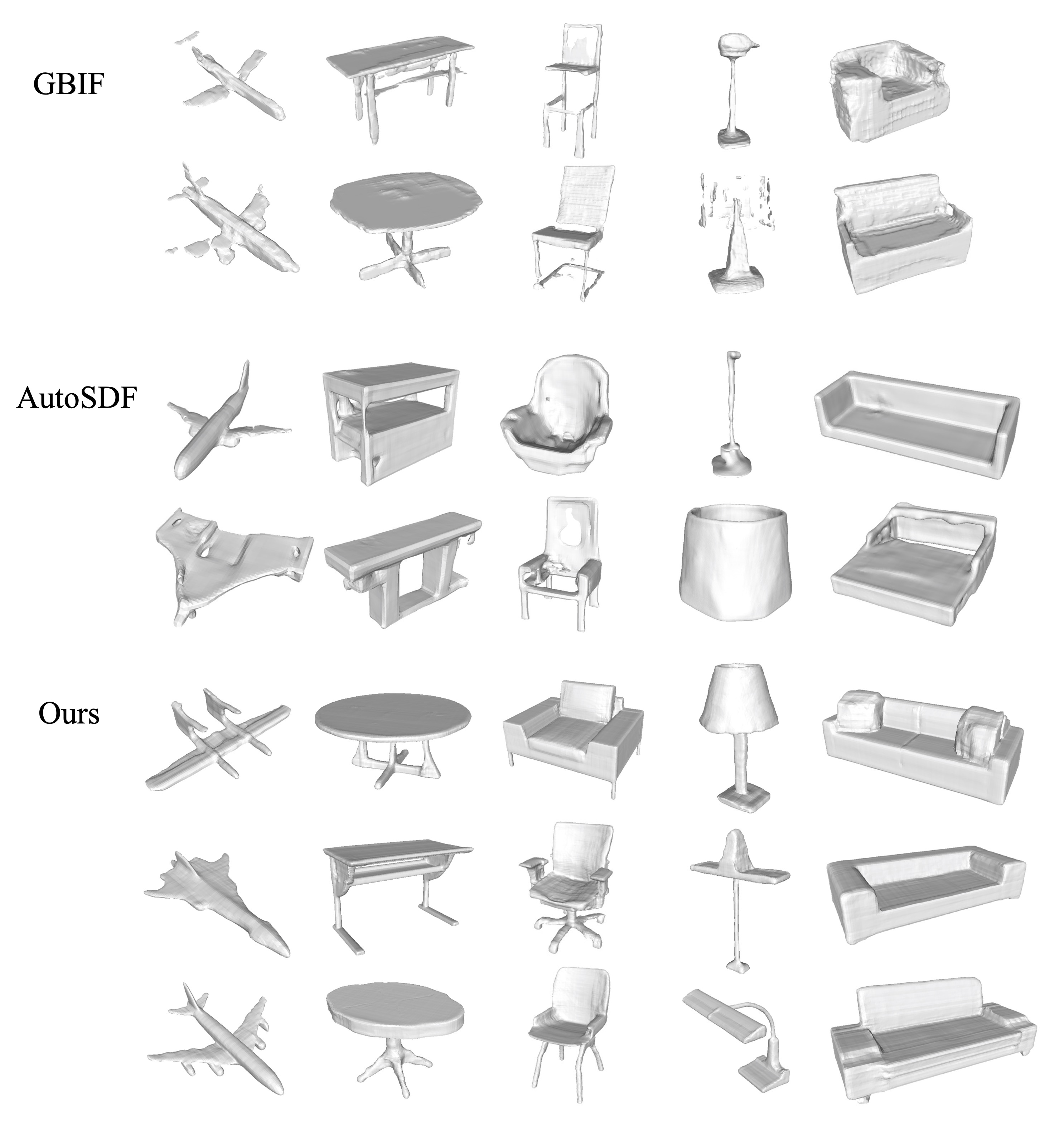}  
\tabularnewline
\vspace{-0.02in}
\caption{Qualitative results of class-guide generation.}
\label{fig:class_cond}
\vspace{-0.2in}
\end{centering}
\end{figure*}

\begin{figure*}[t]
 \begin{centering}
\includegraphics[width=0.9\linewidth]{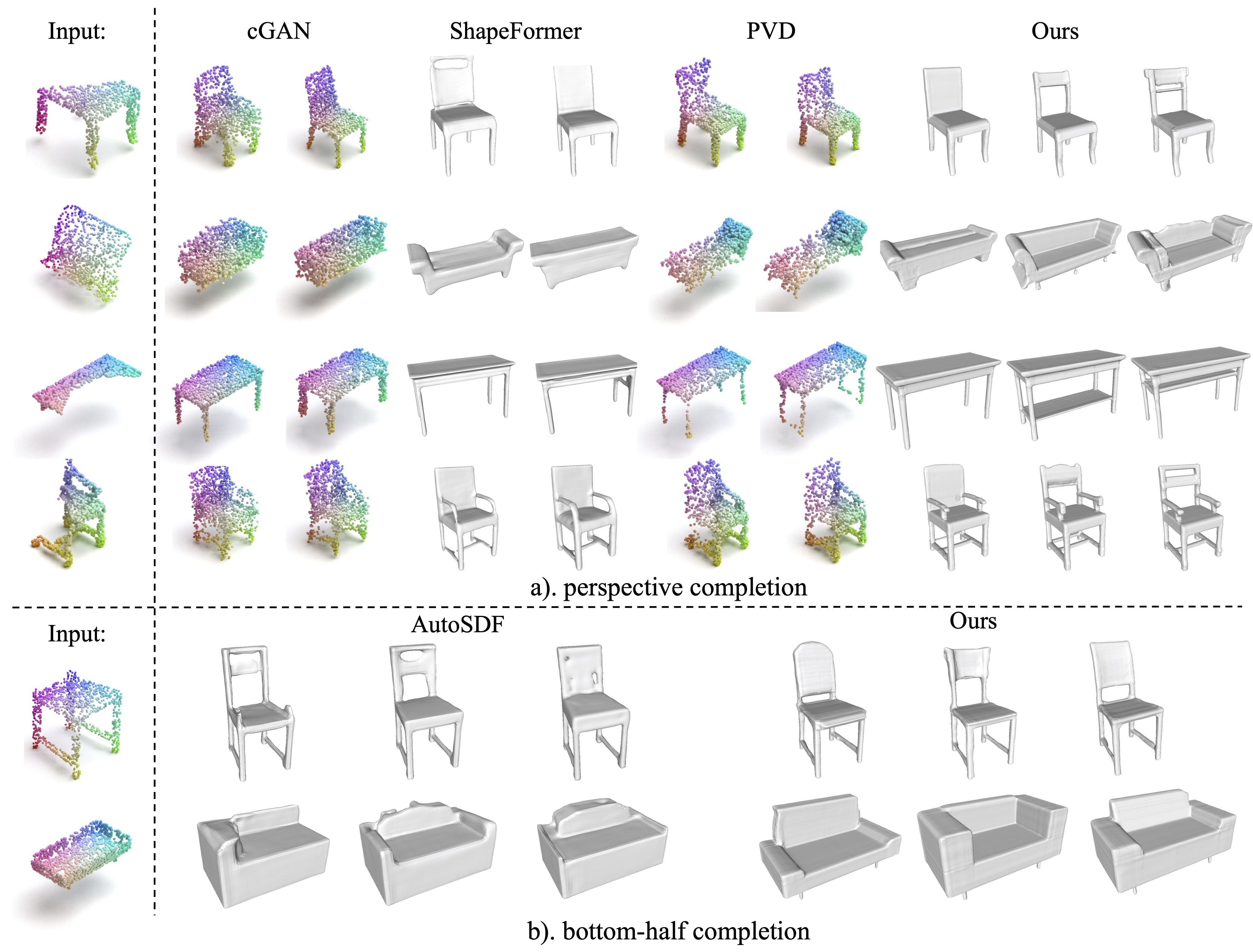}
\vspace{-0.02in}
\caption{Qualitative results of partial point completion. }
\label{fig:shape_completion}
\end{centering}
\end{figure*}

\begin{figure*}[t]
    \centering
    \includegraphics[width=0.9\linewidth]{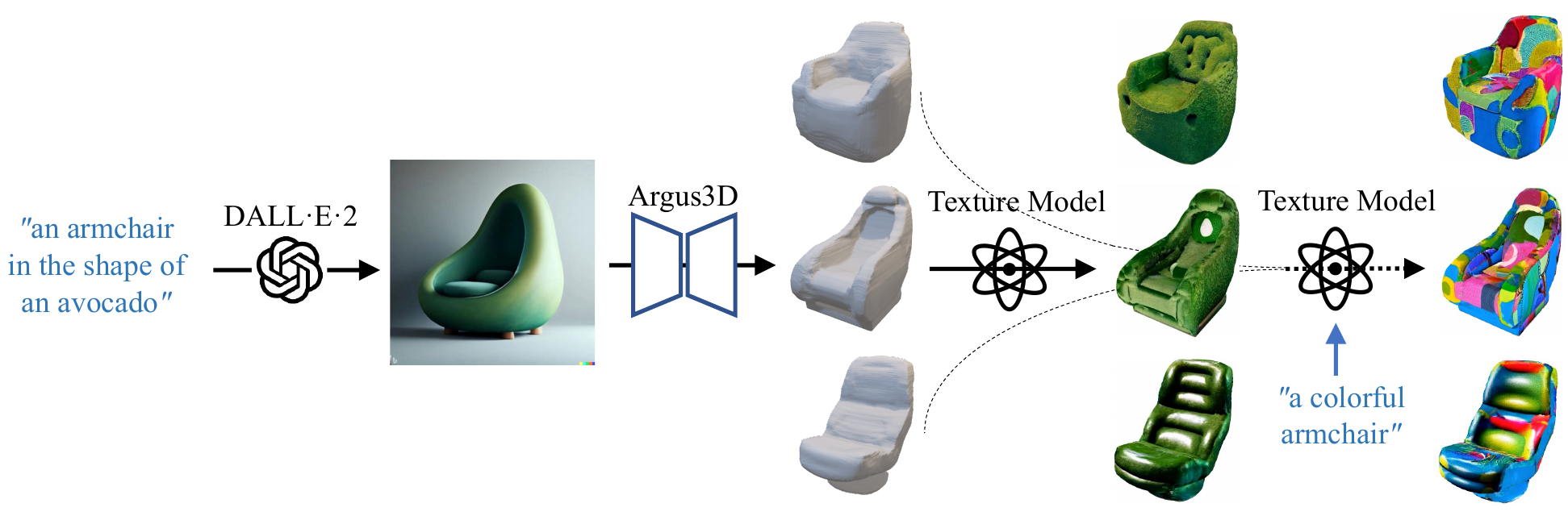}
    \caption{Image-guide generation based on DALL·E·2 generated images. 
    Argus3D is capable of generating a wide range of shapes from unseen images. These shapes can be further enhanced with textures created by a texture model, which utilizes text prompts from DALL·E 2. Additionally, the use of various text prompts enables the generation of new and unique textures.}
    \label{fig:dalle}
\end{figure*}

\end{document}